\documentclass[11pt,letterpaper]{mystyle}

\usepackage[all]{hypcap}
\usepackage{subfigure}
\usepackage{mathtools}
\usepackage{algorithm}
\usepackage{algpseudocode}
     \usepackage{pgfplots}
\pgfplotsset{compat=1.18}
\usepackage{multirow}
\usepackage[capitalize,noabbrev]{cleveref}
\usepackage[textsize=tiny]{todonotes}
\usepackage{fontawesome5}
\usepackage{tcolorbox}
\tcbuselibrary{skins,breakable}
\usepackage{fontawesome5}
\usepackage{enumitem}

\newcommand{\x}{\mathbf{x}}

\definecolor{blanchedalmond}{rgb}{1.0, 0.92, 0.8}
\definecolor{carmine}{rgb}{0.59, 0.0, 0.09}
\definecolor{lightblue}{rgb}{0.22,0.45,0.70}%

\renewcommand{\mathbf}{\boldsymbol}

\makeatletter
\def\Ddots{\mathinner{\mkern1mu\raise\p@
\vbox{\kern7\p@\hbox{.}}\mkern2mu
\raise4\p@\hbox{.}\mkern2mu\raise7\p@\hbox{.}\mkern1mu}}
\makeatother

\definecolor{amaranth}{rgb}{0.9, 0.17, 0.31}
\definecolor{antiquebrass}{rgb}{0.8, 0.58, 0.46}
\definecolor{antiquefuchsia}{rgb}{0.57, 0.36, 0.51}
\definecolor{chromeyellow}{rgb}{0.31, 0.47, 0.26}

\newcommand{\cmark}{\textcolor{green!60!black}{\ding{51}}}
\newcommand{\xmark}{\textcolor{red!70!black}{\ding{55}}}

\title{\textcolor{LynxColor}{LYNX~\faCat}: Learning Dynamic Exits for Confidence-Controlled Reasoning}
\runningtitle{LYNX~\faCat: Learning Dynamic Exits for Confidence-Controlled Reasoning}

\author{
  \textbf{Ömer Faruk Akgül}\textsuperscript{*,1},
  \textbf{Yusuf Hakan Kalaycı}\textsuperscript{*,1},\\
  \vspace{-0.8em}%
  \textbf{Rajgopal Kannan}\textsuperscript{2},
  \textbf{Willie Neiswanger}\textsuperscript{1},
  \textbf{Viktor Prasanna}\textsuperscript{1}\\
  \vspace{0.1em}
  {\normalsize \textsuperscript{1}University of Southern California, \textsuperscript{2}DEVCOM ARL}
}

\correspondingauthor{Ömer Faruk Akgül \href{mailto:}{akgul@usc.edu}. *Equal contribution.}

\begin{document}

\begin{abstract}
\vspace{0.7em}
Large reasoning models achieve strong performance on complex tasks by generating extended chains of thought, but they often ``overthink'': continuing to reason long after they internally have enough information to answer correctly. This wastes inference-time compute and can even hurt accuracy. Existing attempts to stop early either manipulate decoding with extra sampling and heuristics, rely on auxiliary verifier models, or operate only as post-hoc analysis pipelines without formal guarantees. We introduce \emph{LYNX}, an online early-exit mechanism that turns a model's own hidden-state awareness into confidence-controlled stopping decisions. LYNX attaches exit decisions to naturally occurring reasoning cues (e.g., ``hmm'', ``wait'') during generation, trains a lightweight probe on hidden states at those cue tokens using supervision from \emph{forced exits}, and wraps the resulting scores in split conformal prediction to obtain distribution-free control over the rate of premature exits. Crucially, we train and calibrate this probe once on a generic mathematical corpus and then reuse it unchanged across benchmarks, decoding temperatures, and even non-mathematical tasks. Across three model families spanning 1.5B to 32B parameters (DeepSeek-R1-1.5B, QwQ-32B, and Llama-3.1-Nemotron-8B), a single mathematically trained probe per base model yields strong accuracy--efficiency tradeoffs. On \textsc{GSM8K}, LYNX matches or improves baseline accuracy while reducing tokens by $40$--$65\%$; on \textsc{MATH-500} it improves accuracy by up to $12$ points with roughly $35$--$60\%$ fewer tokens; on \textsc{AIME 2024} it recovers baseline accuracy with more than $50\%$ token savings; and on \textsc{CommonsenseQA}, a non-math benchmark, it transfers zero-shot with modest accuracy gains and up to $70\%$ fewer tokens. Compared to state-of-the-art early-exit methods, LYNX offers competitive or superior Pareto frontiers while remaining fully online, requiring no proxy models at inference, and providing explicit, user-tunable confidence guarantees. Code is available at \url{https://github.com/farukakgul/LYNX}.
\end{abstract}

\maketitle

\vspace{-2mm}

\begin{figure}[h!]
    \centering
    \includegraphics[width=\textwidth]{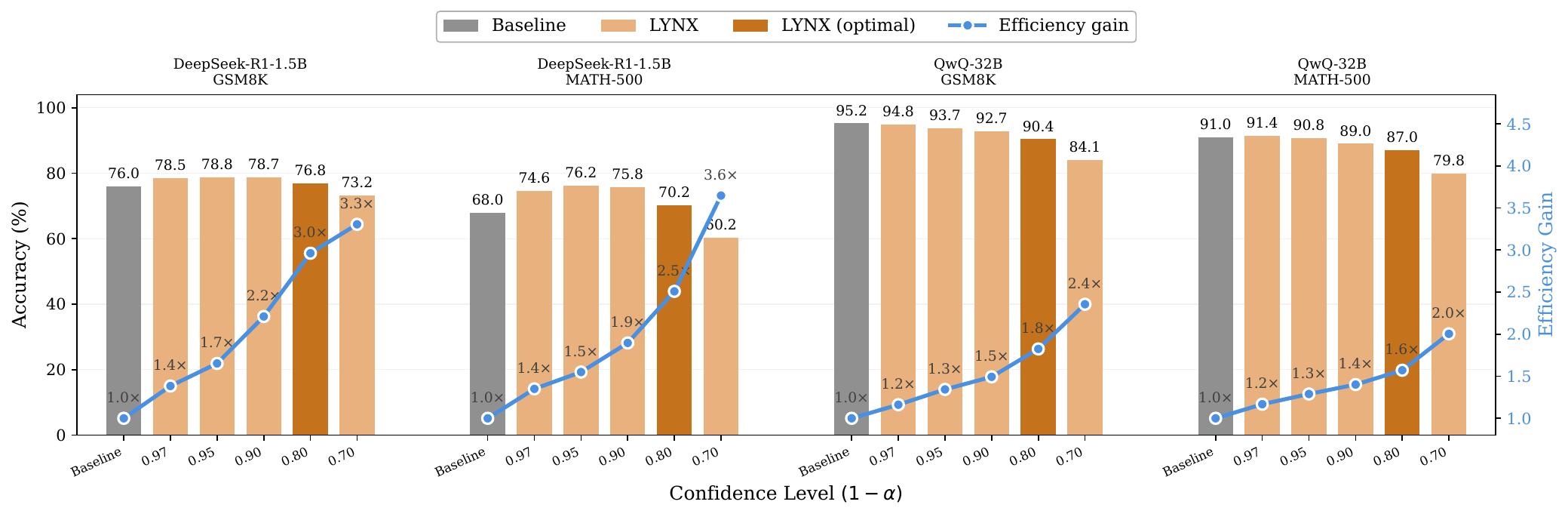}
\caption{Accuracy--efficiency tradeoffs under LYNX: a single confidence parameter smoothly moves the model from baseline-level accuracy to substantially more efficient generations (1.5--3.3$\times$).}
    \vspace{-8mm}
    \label{fig:lynx-confidence-levels}
\end{figure}

    \label{fig:lynx-confidence-levels}

\vspace{-8mm}
\section{Introduction}
Reasoning-oriented large language models (LLMs) have recently achieved breakthrough performance on mathematical problem solving, logical puzzles, and competitive programming by generating long, explicit chains of thought \citep{wei2022chain,openai2024o1}. Models such as OpenAI's o1 \citep{openai2024o1}, DeepSeek-R1 \citep{guo2025deepseekr1}, Google DeepMind's Gemini 2.5 \citep{googledeepmindsearch2025gemini25}, and Anthropic's Claude 3.7 Sonnet \citep{anthropic2025claude37} routinely emit hundreds or thousands of tokens per problem, exploring and revising candidate solution paths before committing to a final answer. This test-time ``thinking'' is a major source of their strength, but it also raises a central question: \emph{how can we allocate this inference-time compute more intelligently, stopping early when further reasoning is unlikely to help?}

Recent empirical studies reveal that large reasoning models frequently exhibit three critical failure modes during extended chain-of-thought generation \citep{su2025between}. First, models often arrive at correct intermediate solutions but continue reasoning and talk themselves into incorrect final answers, a form of counterproductive overthinking. Second, even when maintaining correctness, models generate extensive redundant reasoning that provides no additional value, wasting computational resources without improving accuracy. Third, some problems trigger unbounded reasoning loops that exhaust the model's context window before producing any answer, leading to complete failure. These observations suggest that intelligent early-exit mechanisms are not merely an efficiency optimization but a fundamental requirement for reliable deployment of reasoning models.

In this paper we ask: \emph{Can we turn a model's internal confidence signals into an \textbf{online}, cue-triggered early-exit rule with explicit statistical guarantees?} We answer this in the affirmative with \emph{LYNX}, a framework that enables reasoning models to leverage their own hidden-state awareness for confidence-controlled stopping decisions. Unlike existing approaches that manipulate decoding or rely on external verifier models, LYNX operates along three key axes:

\begin{enumerate}
    \item \textbf{Cue-triggered exits.} Reasoning models naturally emit structural cues during their thinking process, tokens like ``hmm'', ``wait'', ``alternatively'', or paragraph breaks that mark transitions between reasoning segments. Rather than suppressing these cues, we treat them as \emph{decision points}: whenever a cue appears, we consider stopping.
    \item \textbf{Self-contained supervision.} Instead of relying on a proxy LLM to extract and grade intermediate answers, we generate training labels by \emph{forcing exits} at cue positions during data collection: we cut the chain of thought, prompt the base model to output its final answer, and compare against the ground-truth label. This yields a large set of cue-level examples where each hidden state is labeled as ``safe to exit now'' or ``continue thinking'', using only the base model and task answers.
    \item \textbf{Conformal confidence control.} A fixed probe threshold is brittle across datasets and decoding settings. We therefore wrap probe scores in split conformal prediction \citep{vovk2005algorithmic,shafer2008tutorial,angelopoulos2021gentle}, turning them into an adaptive stop/continue rule with a user-chosen miscoverage level. This yields a distribution-free guarantee on the rate of premature exits under mild exchangeability assumptions.
\end{enumerate}

A central design goal of LYNX is \emph{deployability}. Training a new probe and retuning thresholds for every benchmark would undermine the goal of efficient inference. Instead, for each base model, we train and calibrate LYNX \emph{once} on a generic mathematical corpus and then freeze it. In our experiments, we instantiate LYNX on modern open-source reasoning models from three distinct families, including Qwen-based architectures (DeepSeek-R1-Distill-Qwen-1.5B, QwQ-32B) and Llama-based architectures (Llama-3.1-Nemotron-8B), using a moderate number of problems from a mathematical training corpus to (i) collect cue-triggered forced-exit labels and (ii) calibrate conformal thresholds. The resulting probe and thresholds are then applied \emph{unchanged} across four math benchmarks and a non-math benchmark, demonstrating strong generalization across model scales (1.5B to 32B parameters), architectural families, task domains, and decoding temperatures.

Our experiments show that this simple, reusable early-exit head yields strong accuracy–efficiency tradeoffs. On \textsc{GSM8K}, LYNX matches or improves baseline accuracy while reducing tokens by up to $60$--$65\%$ across different models; on \textsc{MATH-500}, it improves accuracy by up to $12$ points with approximately $35$--$60\%$ fewer tokens; on \textsc{AIME 2024}, it recovers baseline accuracy with more than half the tokens removed; and on \textsc{CommonsenseQA} it transfers zero-shot with modest accuracy gains and substantial token savings. When compared to state-of-the-art early-exit methods under matched settings, LYNX consistently provides more favorable Pareto frontiers, while remaining fully online and providing explicit confidence guarantees.

In summary, this paper makes the following contributions:
\begin{enumerate}
    \item We introduce \textbf{\textcolor{LynxColor}{LYNX~\faCat}}, a novel online early-exit framework that combines cue-triggered decision points, self-contained probe training via forced exits, and conformalized confidence control to enable reasoning models to make statistically-guaranteed stopping decisions based on their own hidden states, without modifying decoding, injecting tokens, or requiring external verifiers.
    \item We demonstrate \textbf{cross-architecture generalizability} across three model families (Qwen and Llama), model scales (1.5B to 32B parameters), and task domains (mathematical and commonsense reasoning), showing that a single mathematically-trained probe per base model transfers effectively without per-task retraining.
    \item We establish \textbf{superior accuracy–efficiency tradeoffs} compared to state-of-the-art methods, with experiments showing LYNX matches or improves baseline accuracy while achieving $35$--$65\%$ token reductions, and provide the first deployment-ready early-exit method with explicit, user-tunable confidence guarantees via conformal prediction.
\end{enumerate}

\section{Related Work}
\label{sec:related}
We situate LYNX within three complementary research areas: probing neural representations, inference-time optimization for reasoning, and conformal prediction for reliable decisions.

\vspace{-6mm}

\paragraph{Probing internal representations for verification.}
Probing uses simple classifiers trained on hidden states to extract information that models encode internally but may not explicitly output \citep{alain2016understanding,hewitt2019designing}. Zhang et al.\ demonstrate that reasoning models encode rich signals about solution correctness in their hidden states: a lightweight probe on activations at intermediate ``final answer'' tokens can predict answer correctness with strong calibration on mathematical benchmarks \citep{zhang2025reasoning}. Related work on latent knowledge shows that hidden states contain information about truth and uncertainty even when not directly expressed in outputs \citep{burns2023discovering}. However, these verification pipelines are fundamentally post-hoc, requiring full trace generation, answer extraction using auxiliary models, offline labeling, and post-processing. LYNX bridges this gap by using probes \emph{online} during generation: we train on hidden states at naturally occurring cues using self-contained supervision from forced exits, enabling confidence-controlled stopping decisions without external labeling.
\vspace{-6mm}

\paragraph{Inference-time optimization for efficient reasoning.}
Large reasoning models frequently overthink, generating unnecessarily long chains of thought that waste compute and sometimes hurt accuracy \citep{zhang2025reasoning,yang2025dynamic,jiang2025flashthink,yong2025think,tan2025zerostep,wang2025wait,su2025between}. DEER \citep{yang2025dynamic} detects action-transition tokens as exit points, elicits trial answers, and uses entropy-based confidence scores over token probabilities to decide whether to stop. Think-or-Not \citep{yong2025think} frames reasoning length as mode selection, using tree search over candidate continuations to estimate confidence. However, its efficiency gains can be misleading, as it performs speculative sampling during generation but reports only visible output tokens, hiding substantial inference costs. Other approaches include FlashThink \citep{jiang2025flashthink}, which queries external verifiers on partial reasoning; NoWait \citep{wang2025wait}, which suppresses self-reflection tokens; Dynasor-CoT \citep{fu2025reasoning}, which monitors internal uncertainty signals; and Zero-Step Thinking \citep{tan2025zerostep}, which routes between reasoning modes. While these methods achieve token savings, they typically manipulate generation through extra prompting, verifier queries, or speculative sampling; rely on hand-tuned thresholds requiring recalibration across models and temperatures; and provide no formal statistical control over incorrect early exits. By contrast, LYNX leaves decoding unchanged, operates solely on hidden states at natural cues, requires no auxiliary models at inference, and provides distribution-free guarantees via conformalized probe scores.

\begin{table}[t]
\centering
\vspace{-1mm}
\caption{Qualitative comparison of recent efficient-reasoning methods.
``Online'' indicates whether the method can decide to stop during generation; ``No proxy LLM'' whether inference uses only the base model; ``Self-contained data'' whether training labels can be obtained without external labelers; ``Stat. guarantee'' whether the method provides a distribution-free guarantee on early exit decisions; ``Cue-triggered'' whether the method attaches decisions to naturally occurring reasoning cues; and ``Multi-bench.'' whether a single configuration is reused across multiple benchmarks without per-task tuning.}
\label{tab:method-comparison}
\small
\setlength{\tabcolsep}{4.5pt}
\begin{tabularx}{\linewidth}{@{}lcccccc@{}}
\toprule
\textbf{Method} &
\rotatebox{49}{Online / in-loop} &
\rotatebox{49}{No proxy LLM (inf.)} &
\rotatebox{49}{Self-contained data} &
\rotatebox{49}{Stat. guarantee} &
\rotatebox{49}{Cue-triggered} &
\rotatebox{49}{Multi-bench.} \\
\midrule
\textbf{\textcolor{LynxColor}{LYNX~\faCat}} & \cmark & \cmark & \cmark & \cmark & \cmark & \cmark \\
DEER \citep{yang2025dynamic} & \cmark & \cmark & \cmark & \xmark & \cmark & \xmark \\
Dynasor-CoT \citep{fu2025reasoning} & \cmark & \cmark & \xmark & \xmark & \cmark & \cmark \\
NoWait \citep{wang2025wait} & \cmark & \cmark & \cmark & \xmark & \xmark & \cmark \\
FlashThink \citep{jiang2025flashthink} & \cmark & \xmark & \xmark & \xmark & \xmark & \cmark \\
Think-or-Not \citep{yong2025think} & \cmark & \cmark & \xmark & \xmark & \xmark & \xmark \\
Zhang et al.\ \citep{zhang2025reasoning} & \xmark & \xmark & \xmark & \xmark & \xmark & \xmark \\
\bottomrule
\end{tabularx}
\vspace{-1mm}
\end{table}

\vspace{-6mm}

\paragraph{Conformal prediction for reliable decisions.}
Conformal prediction provides finite-sample, distribution-free guarantees on prediction error, requiring only that test examples are exchangeable with a held-out calibration set \citep{vovk2005algorithmic,shafer2008tutorial,angelopoulos2021gentle}. Recent work applies conformal methods to language models: Quach et al.\ construct prediction sets over token sequences with coverage guarantees \citep{quach2023conformal}; and Savcisens et al.\ uses conformal probes to analyze the trilemma between accuracy, informativeness, and guaranteed coverage in LLM outputs \citep{savcisens2025trilemma}. In the standard setting, given a base predictor $f$, calibration set $\{(x_i, y_i)\}_{i=1}^n$, and nonconformity score $S(x, y)$, one computes the empirical $(1-\delta)$-quantile $q_{1-\delta}$ of calibration scores to form prediction sets $\mathcal{C}(x_{\text{test}}) = \{y : S(x_{\text{test}}, y) \le q_{1-\delta}\}$ that satisfy $\Pr[y_{\text{test}} \in \mathcal{C}(x_{\text{test}})] \ge 1 - \delta$. LYNX applies this framework to probe scores estimating whether a partial chain of thought is sufficient to answer correctly, yielding a conformalized threshold that implements a stop/continue rule with user-chosen confidence level $c = 1 - \delta$ and an explicit guarantee on incorrect early exits. To our knowledge, LYNX is the first early-exit method for reasoning models to provide such distribution-free statistical guarantees.


\section{Method}
\label{sec:method}

\begin{figure}[t]
    \centering
    \includegraphics[width=\linewidth]{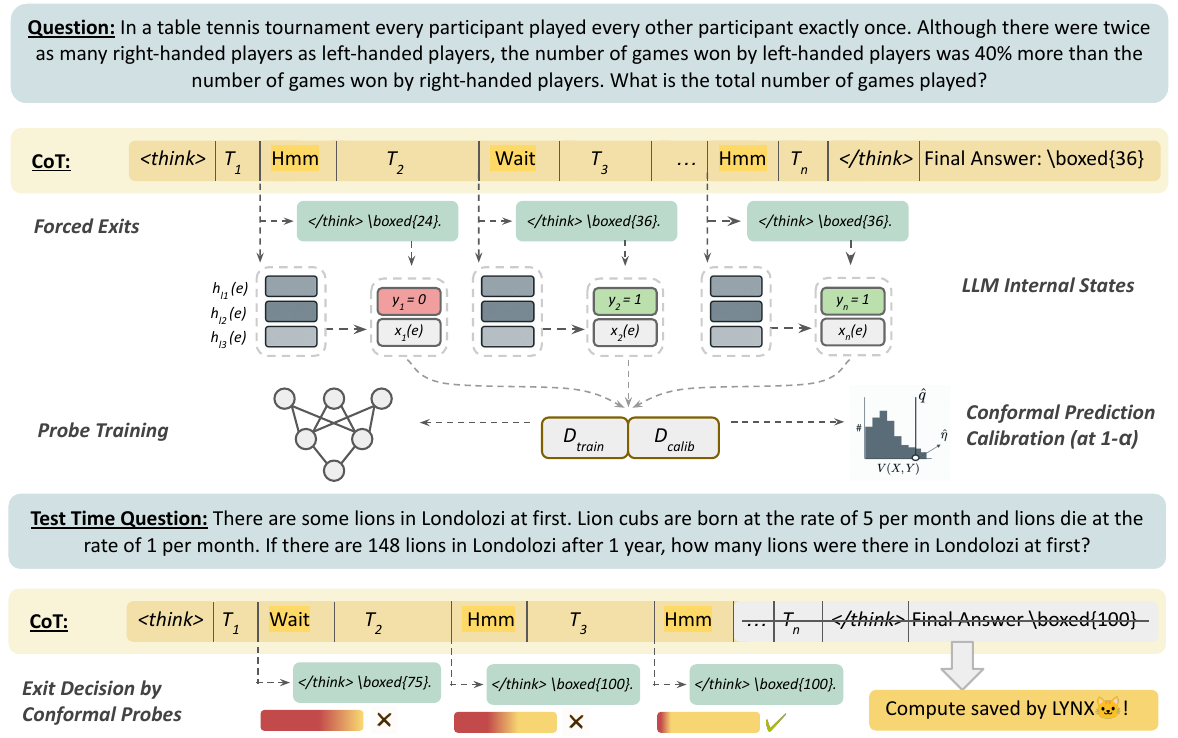}
    \caption{LYNX pipeline: \textbf{(Offline)} We collect forced-exit labels at naturally occurring cue tokens from mathematical problems, train a lightweight probe on hidden states at these cues, and calibrate conformal thresholds using a held-out set. \textbf{(Online)} During generation, whenever a cue appears, the probe scores its hidden state and a conformalized threshold decides whether to exit early or continue reasoning.}
    \label{fig:lynx-overview}
\end{figure}

Having established that (i) reasoning models encode internal correctness signals in hidden states, (ii) existing early-exit methods lack statistical guarantees and require per-task tuning, and (iii) conformal prediction can provide distribution-free control over prediction errors, we now describe how LYNX combines these insights into a practical early-exit framework. The key challenge is to transform post-hoc probing pipelines into an online decision mechanism with explicit confidence guarantees that generalizes across tasks without retraining.

LYNX operates in two stages (Figure~\ref{fig:lynx-overview}). Offline, we generate chains-of-thought on a mathematical corpus, identify natural cue tokens, construct forced exits at each cue to obtain supervision, and train a probe with conformal calibration on hidden states. Online, we run the base model unchanged; whenever a cue appears, the probe scores its hidden state, and a conformalized threshold decides whether to exit or continue. We formalize each component below, and provide a consolidated pseudo-code description in Algorithm~\ref{alg:lynx} in Appendix~\ref{app:algorithm}.

\subsection{Reasoning model and cue tokens}
\label{sec:setup}

Let $x$ be a problem and let the autoregressive LLM $f_\theta$ generate a sequence of tokens
\begin{equation}
(x, t_1, t_2, \dots, t_T) = f_\theta(x),
\end{equation}
with an explicit ``thinking'' segment typically delimited by special tokens such as \texttt{\textless think\textgreater} and \texttt{\textless/think\textgreater}. For each position $j$, we denote by $h_\ell(j) \in \mathbb{R}^d$ the hidden state of layer $\ell$ at token $t_j$. We write $L$ for the total number of layers.

Reasoning models like DeepSeek-R1 often emit self-reflective tokens, ``hmm'', ``wait'', sentence breaks, etc., that mark transitions between reasoning segments (e.g., switching from exploration to verification, or revising a previous plan). We treat these as natural decision points and define a small set of cue token types
\begin{equation}
\mathcal{S} = \{\text{``hmm''},\ \text{``wait''},\ \text{``alternatively''}\},
\end{equation}
and for a given trace we collect all positions
\begin{equation}
E(x) = \{ e \in \{1,\dots,T\} : t_e \in \mathcal{S} \}.
\end{equation}
These cue positions are the only places where LYNX considers early exit. Between cues, generation proceeds exactly as in standard CoT decoding.

\subsection{Supervision via forced exits}
\label{sec:forced-exits}

We assume a supervised training corpus
$\mathcal{D}_{\text{train}} = \{(x_i, a_i^\star)\}_{i=1}^N$,
where $a_i^\star$ is the gold answer for problem $x_i$.
For each $(x_i, a_i^\star)$, we perform a single long CoT rollout with the base model under a fixed decoding configuration (temperature, max length, etc.), obtaining tokens $\{t_{i,j}\}_{j=1}^{T_i}$ and hidden states $\{h_{i,\ell}(j)\}_{\ell=1}^L$.
We then extract all cue positions
$E_i = \{ e \in \{1,\dots,T_i\} : t_{i,e} \in \mathcal{S}\}$.

For every cue $e \in E_i$, we construct a \emph{counterfactual forced exit}. Let $\mathrm{pref}_i(e)$ denote the prompt consisting of $x_i$ and the generated tokens up to (but not including) $t_{i,e}$. We append the model's native answer cue (e.g.,
\texttt{\textless/think\textgreater\ Final Answer: \textbackslash boxed\{})
and let the base model generate a short answer-only continuation:
\[
\hat{a}_i(e) = \mathrm{Exit}_\theta(x_i, e).
\]
We parse this continuation into a normalized answer $\mathrm{parse}(\hat{a}_i(e))$ using the same rule-based extractor as in evaluation, and define a binary label
\begin{equation}
y_i(e) = \mathbf{1}\big[ \mathrm{parse}(\hat{a}_i(e)) = a_i^\star \big] \in \{0,1\}.
\end{equation}

This procedure yields a dataset of cue-level examples
$\mathcal{D}_{\text{cue}} = \big\{(x_i, e, y_i(e)) : 1 \le i \le N,\ e \in E_i \big\}$,
where each example answers: ``If we had exited at this cue, would the resulting answer have been correct?'' Crucially, labels are obtained using only the base model and task answers; no external LLM or human annotation is required.

We partition $\mathcal{D}_{\text{cue}}$ into a probe-training subset $\mathcal{D}_{\text{probe}}$ and a calibration subset $\mathcal{D}_{\text{cal}}$, which will be used for conformal threshold estimation.

\subsection{Probe representation and training}
\label{sec:probe}

At each cue $(x_i, e)$ we turn the corresponding hidden states into a fixed-dimensional representation. Let
$h_{i,\ell}(e) \in \mathbb{R}^d$ denote the hidden state at token $t_{i,e}$ in layer $\ell$.
Rather than relying on a single layer, we aggregate information from a small set of layers
$\mathcal{L}_\text{sel} = \{\ell_1, \ell_2, L\}$ (two middle layers and the final layer), and define
\begin{equation}
z_i(e) = \big[ h_{i,\ell_1}(e) \,\Vert\, h_{i,\ell_2}(e) \,\Vert\, h_{i,L}(e) \big] \in \mathbb{R}^{d'},
\end{equation}
where $\Vert$ denotes concatenation.
Recent empirical studies have shown that different layers of LLMs encode reasoning patterns at varying levels of granularity, ranging from local token-level relationships to task-specific predictions. \citep{jawahar2019does,van2019does}.

We then train a small multilayer perceptron (MLP) probe $g_\psi : \mathbb{R}^{d'} \to [0,1]$ to estimate the probability that an early exit at this cue would be correct:
\begin{equation}
p_i(e) = g_\psi\big(z_i(e)\big) \approx \Pr\big(y_i(e) = 1 \,\big|\, z_i(e)\big).
\end{equation}
Concretely, $g_\psi$ is a two-layer MLP with ReLU activations and a sigmoid output:
\begin{align}
u_i(e) &= \mathrm{ReLU}\big(W_1 z_i(e) + b_1\big), \\
p_i(e) &= \sigma\big(W_2 u_i(e) + b_2\big),
\end{align}
with parameters $\psi = (W_1, b_1, W_2, b_2)$.

Since cue-level labels are typically imbalanced, especially for the smaller models, (most forced exits early in reasoning are not yet safe), we use class-imbalance-aware binary cross-entropy. Let $\mathcal{D}_{\text{probe}} = \{(z_k, y_k)\}_{k=1}^{n}$ be the probe-training set, with $n_1$ positive and $n_0$ negative examples. We define class weights
$w_1 = \frac{n}{2 n_1}$, $w_0 = \frac{n}{2 n_0}$,
and train $g_\psi$ by minimizing
\begin{equation}
\mathcal{L}(\psi)
=
- \frac{1}{n}
\sum_{k=1}^n
\Big(
  w_1 y_k \log p_k
  + w_0 (1 - y_k) \log (1 - p_k)
\Big),
\end{equation}
where $p_k = g_\psi(z_k)$.

\subsection{Conformalized early-exit rule}
\label{sec:conformal-rule}

The probe $g_\psi$ produces a scalar confidence score $p_i(e) \in [0,1]$ for each cue, but how should we convert this into an exit decision? A fixed threshold (e.g., ``exit if $p \ge 0.9$'') is brittle across datasets and temperatures. Instead, we apply split conformal prediction (\S\ref{sec:related}) to obtain a data-driven threshold with explicit statistical guarantees.

\paragraph{User-specified confidence and threshold calibration.}
The user selects a target confidence level $c \in (0,1)$, which controls the tradeoff between efficiency and reliability: higher $c$ means more conservative exits (fewer tokens saved, lower error rate), while lower $c$ means more aggressive exits (more tokens saved, higher error rate). We set the miscoverage level $\delta = 1 - c$ and calibrate a threshold $q_\delta$ as follows.

From the calibration subset $\mathcal{D}_{\text{cal}} = \{(z_j, y_j)\}_{j=1}^m$, we compute nonconformity scores for correctly answered cues ($y_j = 1$):
\begin{equation}
s_j = 1 - g_\psi(z_j)
\quad \text{for all } j \text{ with } y_j = 1.
\end{equation}
Intuitively, $s_j$ is small when the probe is confident that exiting at this cue is safe. We then set
\begin{equation}
q_\delta = \mathrm{Quantile}\big(\{s_j : y_j = 1\},\, 1-\delta \big),
\end{equation}
the $(1-\delta)$-quantile of these scores.

\paragraph{Inference-time policy.}
At test time, we generate CoT as usual. Whenever the model emits a cue token at position $e$ for input $x$, we compute $z(x,e)$ and $p(x,e) = g_\psi(z(x,e))$. We then apply the conformalized decision rule:
\begin{equation}
\label{eq:conformal-exit-rule}
\text{Exit at cue } e
\quad\text{iff}\quad
1 - p(x,e) \le q_\delta.
\end{equation}
If the condition fails, we continue decoding until the next cue or the natural end of the CoT.

\paragraph{Statistical guarantee and practical considerations.}
Under the standard exchangeability assumption, that test examples are drawn from the same distribution as calibration data \citep{vovk2005algorithmic,shafer2008tutorial}, this split-conformal construction controls the fraction of incorrect early exits at approximately $\delta$. Formally, if $(x_{\text{test}}, e_{\text{test}})$ is exchangeable with $\mathcal{D}_{\text{cal}}$ and we exit at $e_{\text{test}}$, then $\Pr[y(e_{\text{test}}) = 1] \ge 1 - \delta$ \citep{angelopoulos2021gentle}.

In practice, strict exchangeability would require calibrating on a held-out portion of each test benchmark. However, standard reasoning benchmarks (AIME, AMC etc.) do not provide separate calibration splits, and creating such splits would reduce the number of available test examples. Instead, we calibrate LYNX once on a generic mathematical corpus and reuse the resulting thresholds across all downstream benchmarks. This design choice prioritizes deployability, enabling zero-shot transfer without per-task tuning, at the potential cost of some miscalibration under domain shift. Our experiments (\S\ref{sec:experiments}) show that this single mathematically-trained probe generalizes effectively across multiple math benchmarks and even to non-mathematical commonsense reasoning tasks, suggesting that the probe learns domain-general signals of reasoning sufficiency rather than task-specific patterns. For applications requiring stricter guarantees, practitioners can recalibrate thresholds on a small held-out subset of their target distribution.

For each base model and decoding temperature, we precompute thresholds for a grid of confidence levels (e.g., $c \in \{0.97, 0.95, 0.90, 0.80, 0.70\}$) during the offline calibration stage, allowing users to select their preferred accuracy-efficiency operating point at deployment time without additional computation.

\section{Experiments}
\label{sec:experiments}

Having established LYNX's theoretical framework, we now empirically show how our framework preserves reasoning accuracy while substantially reducing inference cost, with behavior that generalizes across model families, architectures, scales, and tasks.

\subsection{Experimental setup}
\label{sec:exp-setup}

\paragraph{Models and LYNX instantiation.}
We evaluate LYNX on three open-source reasoning models spanning two architectural families and an order of magnitude in scale: \texttt{DeepSeek-R1-Distill-Qwen-1.5B} \citep{deepseekai2025deepseekr1incentivizingreasoningcapability}, \texttt{QwQ-32B} \citep{qwq32b} (both Qwen-based), and \texttt{Llama-3.1-Nemotron-Nano-8B-v1} \citep{bercovich2025llamanemotron} (Llama-based). All three models emit explicit \texttt{<think>} blocks and natural reasoning cues (``hmm'', ``wait'', ``alternatively''), making them suitable for cue-triggered early exit.

For each base model, we instantiate a single LYNX head by training a probe and calibrating conformal thresholds on Hendrycks MATH \citep{hendrycksmath2021} at temperature $T=0.6$, following the procedure in \S\ref{sec:method}. Critically, this same probe is then deployed unchanged at both $T=0.6$ and $T=0.0$ across all evaluation benchmarks, testing LYNX's robustness to decoding temperature shifts without retraining.

\paragraph{Training and calibration data.}
We use Hendrycks MATH \citep{hendrycksmath2021}, a challenging mathematical reasoning dataset with seven subject categories (Algebra, Counting \& Probability, Geometry, Intermediate Algebra, Number Theory, Prealgebra, Precalculus) and five difficulty levels. For the DeepSeek-1.5B model, we draw 400 training and 200 calibration problems per category (2,800 train, 1,400 calibration total); for QwQ-32B and Nemotron-8B, we use 200 train and 100 calibration per category to balance compute constraints. On each problem we generate a single long CoT rollout at $T=0.6$, extract cue positions, construct forced-exit labels as in \S\ref{sec:forced-exits}, and split into probe-training ($\mathcal{D}_{\text{probe}}$) and calibration ($\mathcal{D}_{\text{cal}}$) sets. Implementation details (probe architecture, optimization, hardware) are in Appendix~\ref{app:impl}.

\paragraph{Evaluation benchmarks.}
We evaluate on five benchmarks:
\begin{itemize}[leftmargin=12pt,itemsep=1pt,topsep=2pt]
    \item \textbf{GSM8K} \citep{cobbe2021gsm8k}: Grade-school math word problems (1,319 test examples).
    \item \textbf{MATH-500} \citep{lightman2023lets}: 500-problem subset of competition mathematics.
    \item \textbf{AMC 2023} \citep{amc_23_hf}: American Mathematics Competition problems (40 problems).
    \item \textbf{AIME 2024} \citep{aime_2024_hf}: American Invitational Mathematics Examination (30 problems).
    \item \textbf{CommonsenseQA}: Non-mathematical multiple-choice commonsense reasoning.
\end{itemize}

For mathematical tasks we use standard CoT prompting (``reason step by step, put the final answer in \verb|\boxed{}|'') and parse numeric answers with task-specific normalization. For CommonsenseQA we use CoT-augmented multiple-choice prompts. We evaluate at two temperatures: $T=0.0$ (deterministic, main results) and $T=0.6$ (stochastic, Appendix~\ref{app:temp06}).

\paragraph{Metrics.}
We report pass@1 accuracy (math) or multiple-choice accuracy (CommonsenseQA) and average generated tokens per example, including all thinking and answer tokens. Efficiency is quantified as:
$\Delta\text{Tokens} = \frac{\text{Tokens}_{\text{method}} - \text{Tokens}_{\text{baseline}}}{\text{Tokens}_{\text{baseline}}}$.
We sweep LYNX's confidence parameter $c \in \{0.97, 0.95, 0.90, 0.80, 0.70\}$ to trace accuracy–efficiency curves.

\subsection{Baselines and cost accounting}
\label{sec:baselines}

We compare LYNX against DEER \citep{yang2025dynamic} and Think-or-Not (ToN) \citep{yong2025think}, the current state-of-the-art early-exit methods for reasoning models, chosen for their strong reported performance and public availability. DEER uses entropy-based confidence scores over token probabilities at action-transition tokens to decide whether to stop. ToN performs tree-search over candidate continuations, using token-level entropies to select between long and short reasoning modes. Both methods require hand-tuned thresholds per model and temperature, whereas LYNX provides a single confidence knob $c$ with formal coverage guarantees.

\paragraph{Cost accounting.}
For all methods we count \emph{total generated tokens} by the base model, including any speculative branches or trial answers that do not appear in the final output. For LYNX and DEER this coincides with visible tokens, as our policy never induces extra generations. For ToN we include all internal sampling costs. This provides a realistic proxy for inference cost. We use authors' public implementations with matched prompts, decoding settings, and maximum CoT length (13,000 tokens). All methods are evaluated on the same hardware (Appendix~\ref{app:impl}).

\section{Results}
\label{sec:results}

We now demonstrate that LYNX achieves superior accuracy–efficiency tradeoffs compared to state-of-the-art baselines while providing interpretable, statistically-grounded control over early exits. Our analysis addresses three questions: (i) Does LYNX form better Pareto frontiers than DEER and ToN? (ii) Does the confidence knob $c$ behave predictably across datasets, models, and temperatures? (iii) How do different model families and decoding settings affect early-exit performance?

\subsection{Main results}
\label{sec:main-results-t00}

Table~\ref{tab:main_t00_two_models} summarizes accuracy–efficiency tradeoffs at $T=0.0$ for DeepSeek-R1-1.5B and QwQ-32B across all five benchmarks. For each dataset we report vanilla CoT (Baseline), DEER, ToN, and LYNX at multiple confidence levels $c$. Boldface configurations indicate recommended operating points that balance accuracy preservation with token efficiency: specifically, we highlight (i) the highest-performing baseline method per dataset, and (ii) for LYNX, the configuration that maximizes token savings while maintaining accuracy within 2 percentage points of baseline, a conservative threshold demonstrating that substantial efficiency gains are achievable without sacrificing correctness. These highlighted points serve as interpretable defaults, though users can select any $c$ along LYNX's Pareto frontier depending on their accuracy-vs-efficiency priorities.

\paragraph{Accuracy–efficiency dominance.}
Across GSM8K, MATH-500, and CommonsenseQA, LYNX consistently traces better Pareto frontiers than both baselines. On \textbf{GSM8K} with DeepSeek-1.5B, LYNX at $c=0.80$ \emph{improves} accuracy over baseline (76.80\% vs 75.97\%, +1.09pp) while reducing tokens by 66.2\%. On \textbf{MATH-500}, $c=0.80$ yields +3.24pp accuracy (70.20\% vs 68.00\%) with 60.1\% fewer tokens, simultaneously addressing overthinking and improving correctness. For \textbf{CommonsenseQA}, $c=0.70$ matches baseline accuracy (40.62\% vs 40.54\%) with 73.9\% token reduction, demonstrating zero-shot transfer from mathematical training to non-mathematical reasoning.

\begin{figure}[t]
    \centering
    \includegraphics[width=\linewidth]{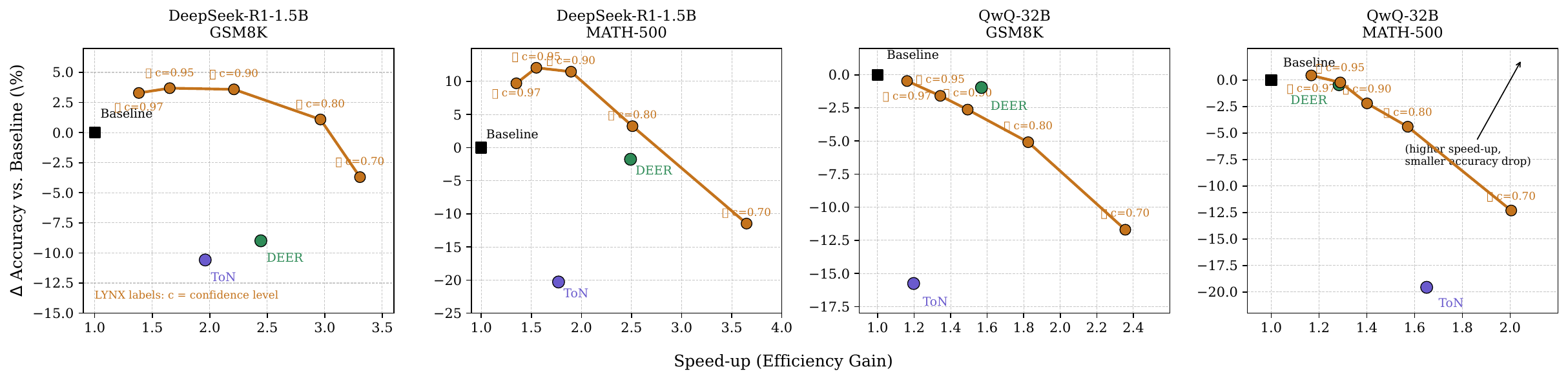}
    \caption{
    Accuracy--efficiency Pareto frontiers for LYNX compared to DEER and Think-or-Not (ToN).
    Each panel plots speed-up (baseline tokens divided by method tokens) on the $x$-axis and change in accuracy vs.\ baseline (percentage points) on the $y$-axis.
    Across all settings, LYNX forms competitive or superior frontiers, with DEER slightly ahead only on QwQ-32B \textsc{GSM8K}.
    }
    \label{fig:pareto-frontiers}
\end{figure}

Similar patterns hold for QwQ-32B: on MATH-500, $c=0.95$ maintains accuracy within 0.22pp while saving 22.4\% of tokens; on CommonsenseQA, $c=0.70$ achieves 61.4\% savings with minimal accuracy loss. Figure~\ref{fig:pareto-frontiers} visualizes these tradeoffs: LYNX curves consistently envelope DEER and ToN operating points across both models on GSM8K and MATH-500.

\paragraph{The confidence knob in practice.}
Although LYNX exposes a continuum of operating points via $c$, users need interpretable defaults. The boldface configurations in Table~\ref{tab:main_t00_two_models} follow a simple rule: maximize token savings subject to with minimal accuracy drop. Under this heuristic, optimal $c$ typically lies in $\{0.80, 0.90, 0.95\}$, saving 32–66\% of tokens across benchmarks. Critically, $c$ behaves monotonically, higher values yield more conservative exits and better accuracy preservation, exactly as predicted by the conformal framework (\S\ref{sec:conformal-rule}). This makes $c$ a meaningful user-facing parameter rather than a brittle hyperparameter requiring per-task grid search.

\paragraph{Comparison to DEER and Think-or-Not.}
DEER and ToN are strong baselines but exhibit characteristic weaknesses. On MATH-500 with DeepSeek-1.5B, DEER achieves 59.8\% token savings but \emph{decreases} accuracy (66.80\% vs 68.00\%), whereas LYNX simultaneously improves accuracy and saves tokens. The exception is AIME 2024, where DEER achieves notable accuracy gains; however, LYNX still provides competitive coverage-controlled operating points (e.g., $c=0.70$ matches baseline accuracy with 55.5\% savings). ToN consistently sits inside LYNX's Pareto frontier once we account for its tree-search overhead, for example, on GSM8K it achieves only 49.0\% savings (vs LYNX's 66.2\% at matched accuracy), despite substantial hidden sampling cost.

\paragraph{Model family and scale generalization.}
Results hold across both Qwen-based models (1.5B and 32B parameters), demonstrating that LYNX scales to larger, more capable reasoning models without retraining. We further validate generalization to the Llama family (Nemotron-8B) in Appendix~\ref{app:nemotron}, where LYNX achieves 70–95\% token reductions on GSM8K and MATH despite this model's distinct architectural characteristics and tendency to use the full 13k token budget (Figure~\ref{fig:nemotron-results}).

\paragraph{Temperature robustness and baseline brittleness.}
A critical advantage of LYNX is deployment at multiple temperatures using a single probe trained at $T=0.6$. Appendix~\ref{app:temp06} shows that LYNX maintains strong performance at $T=0.6$ (e.g., 42.9\% savings on GSM8K at matched accuracy), whereas entropy-based methods like DEER exhibit larger accuracy drops at higher temperatures due to increased output uncertainty. This highlights LYNX's robustness to decoding configuration shifts, a key practical advantage for deployment where temperature may vary by use case.

\definecolor{lynxorange}{HTML}{C4731C}

\newcommand{\lynxmethod}{\textcolor{lynxorange}{\textsc{LYNX}}}

\begin{table*}[t]
\centering
\caption{
Accuracy--efficiency tradeoffs at temperature $T = 0.0$ for DeepSeek-R1-Distill-Qwen-1.5B and QwQ-32B. For each dataset we report vanilla chain-of-thought (\textsc{Baseline}), DEER, Think-or-Not (ToN), and \lynxmethod{} for multiple confidence levels $c$.
$\Delta$Acc vs base and $\Delta$Tokens vs base are relative changes (\%) with respect to each model's own \textsc{Baseline}.
For ToN, the Tokens column reports \emph{total} generated tokens.}
\label{tab:main_t00_two_models}
\scriptsize
\setlength{\tabcolsep}{3pt}
\renewcommand{\arraystretch}{1.05}
\begin{tabular}{l l c
        !{\color{gray}\vrule width 0.8pt}
        cc cc
        !{\color{gray}\vrule width 0.8pt}
        cc cc}
\toprule
 &  &  &
\multicolumn{4}{c}{DeepSeek-R1-1.5B} &
\multicolumn{4}{c}{QwQ-32B} \\
\cmidrule(lr){4-7} \cmidrule(lr){8-11}
Dataset & Method & Conf &
Acc & $\Delta$ vs base & Avg. & $\Delta$ vs base &
Acc & $\Delta$ vs base & Avg. & $\Delta$ vs base \\
\midrule
\multirow{8}{*}{\textsc{GSM8K}} 
  & \textsc{Baseline}      & --   & 0.7597 &   -- & 1751.23 &   -- & 0.9522 &   -- & 1406.12 &   -- \\
\cmidrule(lr){2-11}
  & \textsc{DEER}          & --   & 0.6914 &  -8.99\% &  716.45 & -59.1\%  & \textbf{0.9431} & \textbf{ -0.96\%} &  \textbf{896.41} & \textbf{-36.2\%} \\
  & \textsc{ToN}           & --   & 0.6793 & -10.58\% &  893.04 & -49.0\%  & 0.8021 & -15.76\% &   1174.29 & -16.5\% \\
\cmidrule(lr){2-11}
  & \lynxmethod{}          & 0.97 & 0.7847 &  +3.29\% & 1265.35 & -27.7\%  & 0.9477 &  -0.47\% & 1210.53 & -13.9\% \\
  & \lynxmethod{}          & 0.95 & 0.7877 &  +3.69\% & 1059.88 & -39.5\%  & 0.9371 &  -1.59\% & 1046.63 & -25.6\% \\
  & \lynxmethod{}          & 0.90 & 0.7870 &  +3.59\% &  792.24 & -54.8\%  & 0.9272 &  -2.63\% &  941.72 & -33.0\% \\
  & \textbf{\lynxmethod{}} & \textbf{0.80} & \textbf{0.7680} & \textbf{+1.09\%} & \textbf{591.13} & \textbf{-66.2\%} & 0.9037 &  -5.09\% &  770.66 & -45.2\% \\
  & \lynxmethod{}          & 0.70 & 0.7316 &  -3.70\% &  529.66 & -69.8\%  & 0.8408 & -11.70\% &  596.82 & -57.6\% \\
\midrule
\multirow{8}{*}{\textsc{MATH-500}} 
  & \textsc{Baseline}      & --   & 0.6800 &   -- & 6090.96 &   -- & 0.9100 &   -- & 3945.49 &  -- \\
\cmidrule(lr){2-11}
  & \textsc{DEER}          & --   & 0.6680 &  -1.76\% & 2446.44 & -59.8\%  & 0.9060 &  -0.44\% & 3072.83 & -22.1\% \\
  & \textsc{ToN}           & --   & 0.5420 & -20.29\% & 3438.84 & -43.5\%  & 0.7320 & -19.56\% &  2390.42 & -39.4\% \\
\cmidrule(lr){2-11}
  & \lynxmethod{}          & 0.97 & 0.7460 &  +9.71\% & 4511.43 & -25.9\%  & 0.9140 &  +0.44\% & 3379.23 & -14.4\% \\
  & \lynxmethod{}          & 0.95 & 0.7620 & +12.06\% & 3930.17 & -35.5\%  & \textbf{0.9080} &  \textbf{-0.22}\% & \textbf{3060.71} & \textbf{-22.4\%} \\
  & \lynxmethod{}          & 0.90 & 0.7580 & +11.47\% & 3212.87 & -47.3\%  & 0.8900 &  -2.20\% & 2816.04 & -28.6\% \\
  & \textbf{\lynxmethod{}} & \textbf{0.80} & \textbf{0.7020} & \textbf{+3.24\%} & \textbf{2427.41} & \textbf{-60.1\%} & 0.8700 &  -4.40\% & 2510.93 & -36.4\% \\
  & \lynxmethod{}          & 0.70 & 0.6020 & -11.47\% & 1669.69 & -72.6\%  & 0.7980 & -12.31\% & 1968.30 & -50.1\% \\
\midrule
\multirow{8}{*}{\textsc{AMC 2023}} 
  & \textsc{Baseline}      & --   & 0.5250 &  -- & 9226.08 &  -- & 0.9000 &   -- & 6538.27 &  -- \\
\cmidrule(lr){2-11}
  & \textsc{DEER}          & --   & \textbf{0.5750} &  \textbf{+9.52\%} & \textbf{4846.23} & \textbf{-47.5\%}  & \textbf{0.8750} &  \textbf{-2.78\%} & \textbf{5545.88} & \textbf{-15.2\%} \\
  & \textsc{ToN}           & --   & 0.5000 &  -4.76\% & 5360.97 & -41.9\%  & 0.7000 & -22.22\% & 3680.72 & -43.7\% \\
\cmidrule(lr){2-11}
  & \lynxmethod{}          & 0.97 & 0.5500 &  +4.76\% & 7936.60 & -14.0\%  & 0.8500 &  -5.56\% & 5363.12 & -18.0\% \\
  & \lynxmethod{}          & 0.95 & 0.6250 & +19.05\% & 6865.05 & -25.6\%  & 0.8250 &  -8.33\% & 4828.32 & -26.2\% \\
  & \textbf{\lynxmethod{}} & \textbf{0.90} & \textbf{0.6000} & \textbf{+14.29\%} & \textbf{5315.60} & \textbf{-42.4\%} & \textbf{0.8250} &  \textbf{-8.33\%} & \textbf{4439.62} & \textbf{-32.1\%} \\
  & \lynxmethod{}          & 0.80 & 0.5250 &   0.00\% & 4040.78 & -56.2\%  & 0.7750 & -13.89\% & 3850.82 & -41.1\% \\
  & \lynxmethod{}          & 0.70 & 0.4500 & -14.29\% & 3654.85 & -60.4\%  & 0.6000 & -33.33\% & 2973.38 & -54.5\% \\
\midrule
\multirow{8}{*}{\textsc{AIME 2024}} 
  & \textsc{Baseline}      & --   & 0.2333 &   -- & 13427.43 &   -- & 0.7000 &   -- & 9590.47 &   -- \\
\cmidrule(lr){2-11}
  & \textsc{DEER}          & --   & 0.2000 & -14.27\% &  9154.13 & -31.8\%  & 0.6000 & -14.29\% & 9261.73 &  -3.4\% \\
  & \textsc{ToN}           & --   & 0.2333 &   0.00\% & 11928.16 & -11.2\%  & 0.5333 & -23.81\% & 9612.26 &  +0.2\% \\
\cmidrule(lr){2-11}
  & \lynxmethod{}          & 0.97 & 0.2333 &   0.00\% & 13322.60 &  -0.8\%  & 0.6667 &  -4.76\% & 8440.90 & -12.0\% \\
  & \lynxmethod{}          & 0.95 & 0.2333 &   0.00\% & 13004.03 &  -3.2\%  & 0.6333 &  -9.53\% & 7725.73 & -19.4\% \\
  & \lynxmethod{}          & 0.90 & 0.2000 & -14.27\% & 11635.80 & -13.3\%  & \textbf{0.6333} &  \textbf{-9.53\%} & \textbf{7308.63} & \textbf{-23.8\%} \\
  & \lynxmethod{}          & 0.80 & 0.2333 &   0.00\% &  8501.83 & -36.7\%  & 0.5000 & -28.57\% & 6396.40 & -33.3\% \\
  & \textbf{\lynxmethod{}} & \textbf{0.70} & \textbf{0.2333} & \textbf{0.00\%} & \textbf{5981.83} & \textbf{-55.5\%} & 0.4667 & -33.33\% & 5691.37 & -40.7\% \\
\midrule
\multirow{8}{*}{%
  \begin{tabular}{@{}l@{}}
    \textsc{CommonsenseQA} \\
    (Non-Math)
  \end{tabular}
}
  & \textsc{Baseline}      & --   & 0.4054 &   -- & 2335.69 &   -- & 0.8460 &   -- & 1021.04 &   -- \\
\cmidrule(lr){2-11}
  & \textsc{DEER}          & --   & 0.4136 &  +2.02\% & 1085.95 & -53.5\%  & 0.8493 &  +0.39\% &  622.98 & -39.0\% \\
  & \textsc{ToN}           & --   & \textbf{0.4431} &  \textbf{+9.30}\% &  \textbf{757.52} & \textbf{-67.6\%}  & \textbf{0.8731} &  \textbf{+3.20\%} &    \textbf{543.03} & \textbf{-46.8\%} \\
\cmidrule(lr){2-11}
  & \lynxmethod{}          & 0.97 & 0.4062 &  +0.20\% & 2219.80 &  -5.0\%  & 0.8452 &  -0.09\% &  911.14 & -10.8\% \\
  & \lynxmethod{}          & 0.95 & 0.4038 &  -0.39\% & 2085.18 & -10.7\%  & 0.8485 &  +0.30\% &  770.27 & -24.6\% \\
  & \lynxmethod{}          & 0.90 & 0.4095 &  +1.01\% & 1850.40 & -20.8\%  & 0.8477 &  +0.20\% &  667.02 & -34.7\% \\
  & \lynxmethod{}          & 0.80 & 0.4120 &  +1.63\% & 1254.09 & -46.3\%  & 0.8485 &  +0.30\% &  548.42 & -46.3\% \\
  & \textbf{\lynxmethod{}} & \textbf{0.70} & \textbf{0.4062} & \textbf{+0.20\%} & \textbf{609.70} & \textbf{-73.9\%} & \textbf{0.8444} &  \textbf{-0.19\%} &  \textbf{394.24} & \textbf{-61.4\%} \\
\bottomrule
\end{tabular}
\end{table*}

\subsection{Qualitative analysis: When and why LYNX exits}

Figures~\ref{fig:qualitative-example}–\ref{fig:qualitative-example-books} illustrate LYNX's decision process across four representative failure modes of reasoning models:

\textbf{(1) Redundant verification} (Fig.~\ref{fig:qualitative-example}): The baseline generates 1,105 tokens with extensive re-checking after reaching the correct answer. LYNX exits at the second cue when the conformal predictive set collapses to $\{1\}$, using only 258 tokens (76.6\% reduction) while maintaining correctness.

\textbf{(2) Repetitive reasoning} (Fig.~\ref{fig:qualitative-example2}, Appendix): After solving a multi-step problem correctly, the baseline continues with 193 tokens of redundant verification. LYNX detects sufficiency at cue 6 and exits with 22.1\% savings.

\textbf{(3) Context window exhaustion} (Fig.~\ref{fig:qualitative-example-context}, Appendix): On a challenging MATH problem, the baseline enters circular reasoning and exceeds the 16k token limit \emph{without producing an answer}. LYNX monitors 24 cues, exits when $p_{\text{pos}} \approx 0.999$ at cue 24, and produces the correct answer using only 2,543 tokens, demonstrating that early exit can enable solving problems that would otherwise fail completely.

\textbf{(4) Overthinking to error} (Fig.~\ref{fig:qualitative-example-books}, Appendix): The model correctly calculates the answer but continues reasoning, talks itself out of the correct solution, and outputs an incorrect final answer. LYNX exits \emph{before} this error cascade, preserving the initial correct reasoning.

These examples illustrate that the conformal machinery provides not just aggregate statistical control but also interpretable, cue-level decisions: exits occur only when the calibrated predictive set deems the cue safe at confidence $c = 1 - \delta$.

\begin{figure*}[t]
\centering
\small
\sffamily 

\definecolor{questionblue}{HTML}{E3F2FD}
\definecolor{baselinered}{HTML}{FFEBEE}
\definecolor{baselinedark}{HTML}{C62828}
\definecolor{lynxorange}{HTML}{C4731C}
\definecolor{lynxlight}{HTML}{FFF3E0}
\definecolor{cueyellow}{HTML}{FFD54F}
\definecolor{explaingreen}{HTML}{E8F5E9}

\begin{tcolorbox}[
    colback=questionblue,
    colframe=questionblue!50!black,
    arc=3mm,
    boxrule=0.5pt,
    left=5pt, right=5pt, top=5pt, bottom=5pt,
    width=\textwidth,
    fontupper=\small\sffamily
]
\textbf{Question:} A robe takes 2 bolts of blue fiber and half that much white fiber. How many bolts in total does it take?
\end{tcolorbox}

\vspace{2mm}

\begin{minipage}[t]{0.48\textwidth}
\begin{tcolorbox}[
    colback=baselinered,
    colframe=baselinered!50!black,
    arc=3mm,
    boxrule=0.5pt,
    left=4pt, right=4pt, top=4pt, bottom=4pt,
    fontupper=\footnotesize\sffamily,
    title={\textbf{Baseline (Full Generation)}},
    coltitle=white,
    colbacktitle=baselinedark,
    height=16cm
]
Okay, so I have this problem here: A robe takes 2 bolts of blue fiber and half that much white fiber. How many bolts in total does it take? \colorbox{cueyellow}{\textbf{Hmm}}, let me try to figure this out step by step.

First, I need to understand what the question is asking. It's about figuring out the total number of bolts of fiber needed to make a robe. The robe uses two bolts of blue fiber and half as much white fiber. So, I guess I need to add the number of blue fiber bolts and white fiber bolts together to get the total number of fiber bolts required.

Let me write that down to make it clearer. If the robe takes 2 bolts of blue fiber, then the white fiber is half of that. So, the white fiber would be 2 divided by 2, right? Let me calculate that: 2 divided by 2 is 1. So, the white fiber is 1 bolt.

Now, to find the total number of bolts, I need to add the blue fiber and white fiber together. That would be 2 (blue) plus 1 (white). Adding those together gives me 3. So, the total number of bolts needed is 3.

\colorbox{cueyellow}{\textbf{Wait}}, let me make sure I didn't make a mistake. Sometimes, when dealing with fractions, it's easy to slip up. So, 2 divided by 2 is definitely 1. Yeah, that seems right. So, adding 2 and 1, that gives me 3. So, the total number of bolts is 3.

I think that makes sense. The robe uses twice as much blue fiber as white fiber, so blue is 2 and white is 1. Adding them up gives 3.

\vspace{1mm}
\centering\textcolor{gray}{\textit{... (839 more tokens) ...}}
\vspace{1mm}

\raggedright
\colorbox{cueyellow}{\textbf{Hmm}}, seems like every way I look at it, it's 3. I don't think I can get a different answer. So, I think that's the right answer.

Just to recap: 2 blue, half of that is 1 white, so 2 + 1 = 3. Yeah, that seems solid. [...] I've checked it multiple ways, considered different interpretations, made sure the math adds up, and everything points to 3 being the correct answer.

\vfill

\textbf{Final Answer:} $\boxed{3}$

\vspace{2mm}
\raggedleft\textcolor{gray}{\footnotesize Total tokens: \textbf{1,105}}
\end{tcolorbox}
\end{minipage}
\hfill
\begin{minipage}[t]{0.48\textwidth}
\begin{tcolorbox}[
    colback=lynxlight,
    colframe=lynxorange,
    arc=3mm,
    boxrule=0.5pt,
    left=4pt, right=4pt, top=4pt, bottom=4pt,
    fontupper=\footnotesize\sffamily,
    title={\textbf{LYNX (Early Exit)}},
    coltitle=white,
    colbacktitle=lynxorange,
    height=16cm
]
Okay, so I have this problem here: A robe takes 2 bolts of blue fiber and half that much white fiber. How many bolts in total does it take? \colorbox{cueyellow}{\textbf{Hmm}} \textcolor{lynxorange}{\faTimesCircle} \textcolor{gray}{\footnotesize [Cue 1: $p_{\text{pos}} \approx 0.84$, set = $\{0,1\}$ $\rightarrow$ continue]}, let me try to figure this out step by step.

First, I need to understand what the question is asking. It's about figuring out the total number of bolts of fiber needed to make a robe. The robe uses two bolts of blue fiber and half as much white fiber. So, I guess I need to add the number of blue fiber bolts and white fiber bolts together to get the total number of fiber bolts required.

Let me write that down to make it clearer. If the robe takes 2 bolts of blue fiber, then the white fiber is half of that. So, the white fiber would be 2 divided by 2, right? Let me calculate that: 2 divided by 2 is 1. So, the white fiber is 1 bolt.

Now, to find the total number of bolts, I need to add the blue fiber and white fiber together. That would be 2 (blue) plus 1 (white). Adding those together gives me 3. So, the total number of bolts needed is 3.

\colorbox{cueyellow}{\textbf{Wait}} \textcolor{green!60!black}{\faCheckCircle} \textcolor{gray}{\footnotesize [Cue 2: $p_{\text{pos}} \approx 1.00$, set = $\{1\}$ at $c=0.95$ $\rightarrow$ \textbf{EXIT}]}

\vfill

\textbf{Final Answer:} $\boxed{3}$

\vspace{2mm}
\raggedleft\textcolor{gray}{\footnotesize Total tokens: \textbf{258} \textcolor{green!60!black}{(\textbf{-76.6\% vs baseline})}}

\vspace{3mm}
\tcblower
\begin{tcolorbox}[
    colback=explaingreen,
    colframe=explaingreen!50!black,
    arc=2mm,
    boxrule=0.3pt,
    left=3pt, right=3pt, top=3pt, bottom=3pt,
    fontupper=\footnotesize\sffamily
]
\textbf{Exit Decision Process:}
\begin{itemize}[leftmargin=10pt, itemsep=1pt, topsep=2pt]
    \item \textbf{Cue 1 (``hmm''):} Probe score $p_{\text{pos}} \approx 0.84$. Conformal predictive set: $\{0, 1\}$ (ambiguous) $\Rightarrow$ \textcolor{lynxorange}{\textbf{no exit}}, model continues.
    \item \textbf{Cue 2 (``wait''):} Probe score $p_{\text{pos}} \approx 1.00$. Conformal predictive set: $\{1\}$ at confidence $c = 0.95$ $\Rightarrow$ \textcolor{green!60!black}{\textbf{safe to exit}}.
\end{itemize}
\vspace{1mm}
\textbf{Outcome:} Same correct answer ($\boxed{3}$) as baseline, but LYNX exits at the second cue and uses only the essential part of the chain-of-thought, saving 847 tokens.
\end{tcolorbox}
\end{tcolorbox}
\end{minipage}

\caption{Example outputs comparing baseline generation with LYNX early exit on GSM8K. The baseline model generates 1,105 tokens with extensive overthinking, while LYNX monitors natural reasoning cues (\colorbox{cueyellow}{hmm}, \colorbox{cueyellow}{wait}) and exits confidently after 258 tokens when the conformal predictive set contains only the correct answer. Both arrive at the same correct answer, but LYNX achieves 76.6\% token reduction. The exit is triggered only when the conformal predictive set collapses to $\{1\}$, i.e., when the probe is calibrated to treat the cue as a safe exit at the chosen confidence level.
}
\label{fig:qualitative-example}
\end{figure*}

\section{Conclusion}
\label{sec:conclusion}

We introduced LYNX, a cue-triggered, conformal early-exit mechanism that turns reasoning models' internal hidden-state awareness into confidence-controlled stopping decisions. For each base model, LYNX trains a single lightweight probe using self-contained supervision from forced exits and wraps the resulting scores in split conformal prediction to provide user-tunable guarantees on incorrect early exits. Across three model families (Qwen and Llama), three scales (1.5B–32B parameters), four mathematical benchmarks, and one non-mathematical task, a single mathematically-trained probe per model yields strong accuracy–efficiency tradeoffs: matching or improving baseline accuracy while reducing tokens by 40–95\% depending on the chosen confidence level.

Compared to state-of-the-art methods, LYNX offers competitive or superior Pareto frontiers while requiring no auxiliary models at inference, no decoding modifications, and providing explicit statistical guarantees. The method generalizes across temperatures without retraining and transfers zero-shot to non-mathematical domains, suggesting it captures domain-general signals of reasoning sufficiency rather than task-specific patterns.

\paragraph{Limitations.}
Our study focuses on open-source models with explicit thinking segments and self-reflective cue tokens. Models without such structure may require alternative trigger definitions. Our conformal guarantees assume exchangeability between calibration and deployment data; under strong domain shift the effective coverage may deviate from nominal levels. We train separate heads per base model and temperature-independent probes per family, and do not optimize the cost of forced-exit dataset construction, viewing it as amortized over downstream usage.

\paragraph{Future work.}
Immediate directions include applying LYNX to code generation, multi-modal reasoning, and tool-augmented settings; learning latent decision points beyond hand-specified cues; co-training base models and probes for improved linear readability; and combining LYNX with external verifiers or self-consistency methods for additional reliability gains.

\clearpage
\bibliographystyle{plainnat}
\bibliography{main}

\newpage

\section{Appendix.}
\appendix

\section{Algorithmic Details of \lynxmethod{}}
\label{app:algorithm}

In this section we summarize the full \lynxmethod{} pipeline, including offline cue-level supervision, probe training and conformal calibration, and the online cue-triggered exit policy, in a single algorithm. Algorithm~\ref{alg:lynx} makes explicit the roles of the cue set $\mathcal{S}$, probe-training set $\mathcal{D}_{\text{probe}}$, calibration set $\mathcal{D}_{\text{cal}}$, the probe $g_\psi$, the conformal threshold $q_\delta$, and the user-facing confidence level $c = 1 - \delta$, corresponding to the components described in \S\ref{sec:setup}--\S\ref{sec:conformal-rule}.

\begin{algorithm}[t]
\caption{\lynxmethod{}: Cue-triggered conformal early exit}
\label{alg:lynx}
\begin{algorithmic}[1]
\Require Base model $f_\theta$, training set $\mathcal{D}_{\text{train}} = \{(x_i, a_i^\star)\}$, cue set $\mathcal{S}$, confidence level $c \in (0,1)$
\Ensure Probe $g_\psi$ and conformal threshold $q_\delta$ implementing an early-exit policy at confidence $c = 1 - \delta$
\vspace{0.35ex}
\Statex \textcolor{lynxorange}{\textbf{Offline stage: collect cue-level supervision}}
\For{each $(x_i, a_i^\star) \in \mathcal{D}_{\text{train}}$}
  \State Generate a long chain-of-thought rollout $(t_{i,1},\dots,t_{i,T_i})$ with $f_\theta$ and record hidden states $h_{i,\ell}(j)$
  \State Extract cue positions $E_i = \{ e : t_{i,e} \in \mathcal{S}\}$
  \For{each cue $e \in E_i$}
    \State Form prefix $\mathrm{pref}_i(e)$ consisting of $x_i$ and tokens up to $t_{i,e-1}$
    \State Append the model's answer prompt (e.g., \texttt{\textless/think\textgreater\ Final Answer: \textbackslash boxed\{}), and call
    \[
      \hat{a}_i(e) \gets \mathrm{Exit}_\theta(x_i, e)
    \]
    \State Parse the answer $\tilde{a}_i(e) \gets \mathrm{parse}(\hat{a}_i(e))$
    \State Define label $y_i(e) \gets \mathbf{1}\big[\tilde{a}_i(e) = a_i^\star\big] \in \{0,1\}$
    \State Build representation $z_i(e)$ from hidden states at $e$ (e.g.,
           $z_i(e) = [h_{i,\ell_1}(e) \Vert h_{i,\ell_2}(e) \Vert h_{i,L}(e)]$)
    \State Add $(z_i(e), y_i(e))$ to the cue-level dataset $\mathcal{D}_{\text{cue}}$
  \EndFor
\EndFor
\State Split $\mathcal{D}_{\text{cue}}$ into probe-training set $\mathcal{D}_{\text{probe}}$ and calibration set $\mathcal{D}_{\text{cal}}$
\vspace{0.5ex}
\Statex \textcolor{lynxorange}{\textbf{Offline stage: train probe and calibrate conformal threshold}}
\State Train a small MLP probe $g_\psi : \mathbb{R}^{d'} \to [0,1]$ on $\mathcal{D}_{\text{probe}}$ with class-weighted binary cross-entropy,
  so that $p = g_\psi(z) \approx \Pr[y=1 \mid z]$
\State Set $\delta \gets 1 - c$
\State For all $(z_j, y_j) \in \mathcal{D}_{\text{cal}}$ with $y_j = 1$, compute scores $s_j \gets 1 - g_\psi(z_j)$
\State Let $q_\delta$ be the empirical $(1-\delta)$-quantile of $\{s_j : y_j = 1\}$
\vspace{0.5ex}
\Statex \textcolor{lynxorange}{\textbf{Online stage: conformal early-exit policy}}
\Function{LYNX-Eval}{$x$}
  \State Initialize the prompt with $x$ and start generating a chain-of-thought with $f_\theta$
  \For{each generated token position $e = 1,2,\dots$}
    \If{$t_e \in \mathcal{S}$} \Comment{cue-triggered decision point}
      \State Form representation $z(e)$ from hidden states at $e$
      \State $p \gets g_\psi(z(e))$; \quad $s \gets 1 - p$
      \If{$s \le q_\delta$} \Comment{conformal stop rule at confidence $c$}
        \State Append the model's answer prompt and call $\hat{a} \gets \mathrm{Exit}_\theta(x, e)$
        \State \Return final answer $\mathrm{parse}(\hat{a})$
      \EndIf
    \EndIf
  \EndFor
  \State \Return answer obtained from the full chain-of-thought (no early exit triggered)
\EndFunction
\end{algorithmic}
\end{algorithm}

\section{Implementation Details}
\label{app:impl}

\paragraph{Hardware and compute.}
All experiments run on 4× NVIDIA RTX A6000 (Blackwell, 96GB) GPUs. Maximum CoT length is set to 13,000 tokens for all models and methods.

\paragraph{Probe architecture.}
The probe $g_\psi$ is a two-layer MLP with hidden dimensions $[256, 64]$ and GELU activation, taking as input concatenated hidden states from three layers (two middle, one final) at each cue token. We optimize with Adam (lr=1e-3, weight decay=1e-4), class-weighted binary cross-entropy, and early stopping on a held-out validation split of $\mathcal{D}_{\text{probe}}$.

\paragraph{Training and calibration data sizes.}
For DeepSeek-R1-1.5B: 400 training + 200 calibration problems per Hendrycks MATH category (2,800 train, 1,400 calibration total). For QwQ-32B and Nemotron-8B: 200 train + 100 calibration per category to balance compute (1,400 train, 700 calibration total).

\section{Results at Temperature $T=0.6$}
\label{app:temp06}

Table~\ref{tab:main_t06_two_models} reports accuracy–efficiency tradeoffs at $T=0.6$ for DeepSeek-1.5B and QwQ-32B. At this higher temperature, baseline models exhibit increased stochasticity in generation, which affects entropy-based methods like DEER more severely than LYNX.

\paragraph{Key observations.}
LYNX maintains strong performance at $T=0.6$: on GSM8K with DeepSeek-1.5B, $c=0.90$ achieves near-baseline accuracy (77.86\% vs 78.39\%, -0.68pp) with 42.9\% token savings. On MATH-500, $c=0.95$ preserves accuracy within 2.2pp while reducing tokens by 32.0\%. CommonsenseQA shows 54.6\% savings at matched accuracy with $c=0.70$.

Critically, DEER and ToN exhibit larger accuracy degradation at $T=0.6$ compared to $T=0.0$. On GSM8K, DEER's accuracy drops by 13.83pp at $T=0.6$ vs 8.99pp at $T=0.0$. This suggests entropy-based confidence estimation becomes less reliable under increased sampling stochasticity, whereas LYNX's probe-based approach remains robust. The same mathematically-trained probe deployed at both temperatures demonstrates LYNX's generalization across decoding configurations without per-temperature retuning.

\section{Generalization to Llama-Nemotron}
\label{app:nemotron}

\begin{figure*}[t]
\centering
\includegraphics[width=\textwidth]{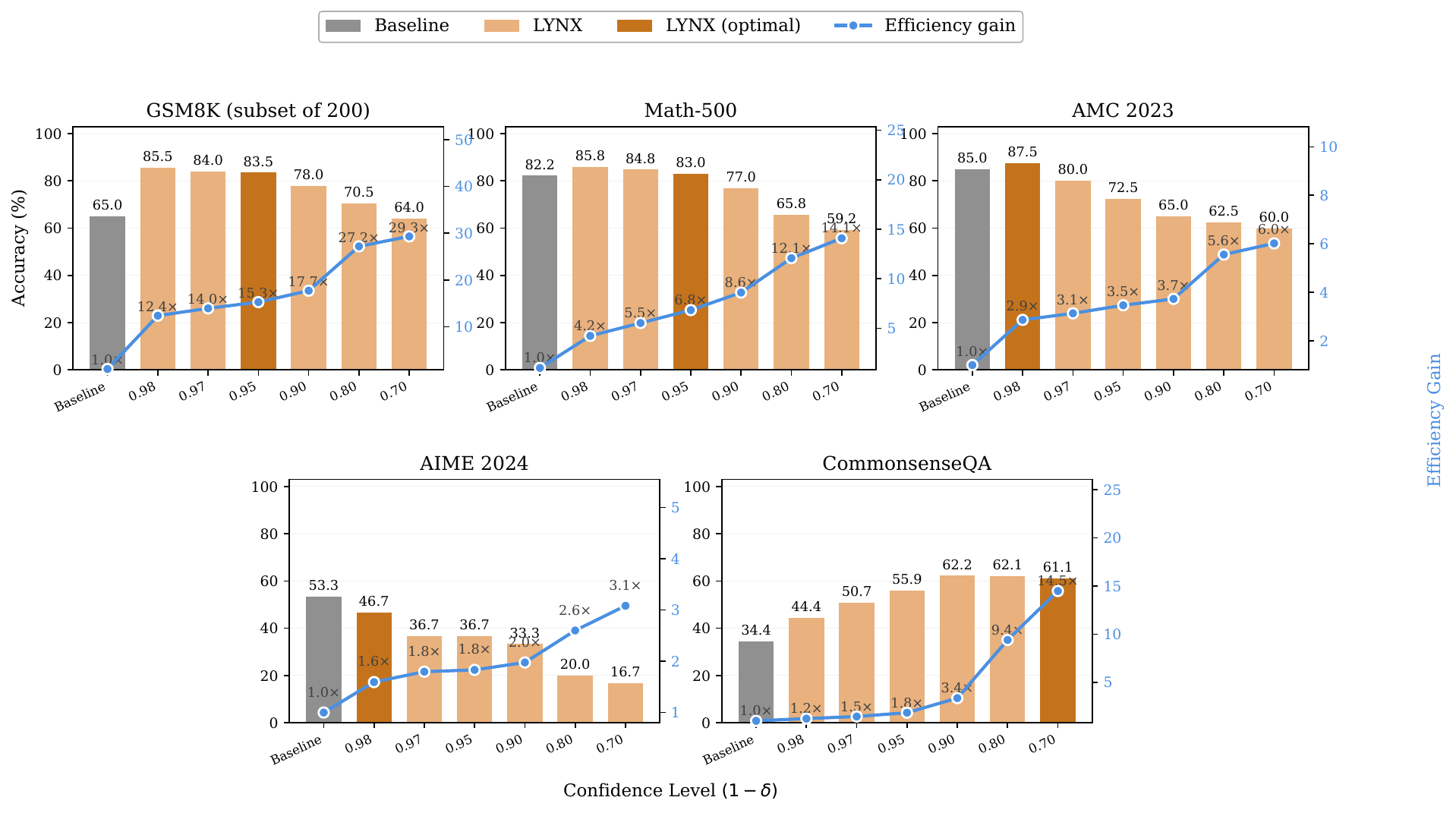}
\caption{
Accuracy–efficiency tradeoffs at temperature $T = 0.0$ for \textsc{Llama-3.1-Nemotron-Nano-8B-v1}.
Each panel shows baseline chain-of-thought decoding and \lynxmethod{} at multiple confidence levels $c = 1 - \delta$.
Bars report accuracy, and the overlaid line reports efficiency gain (baseline tokens divided by method tokens).
\lynxmethod{} achieves large token savings on all datasets while substantially improving or preserving accuracy relative to the Nemotron baseline.}
\label{fig:nemotron-results}
\end{figure*}

Figure~\ref{fig:nemotron-results} reports results for Llama-3.1-Nemotron-Nano-8B-v1 at $T=0.0$.
This model presents a distinct challenge: it consistently uses the full 13k token budget across most problems, suggesting a tendency toward verbose, potentially circular reasoning patterns.
As noted in the DEER paper, Nemotron exhibits lower confidence in generating intermediate answers, resulting in significantly lower early-stopping rates compared to Qwen-based models.

\paragraph{LYNX performance.}
Despite this challenging profile, \lynxmethod{} achieves dramatic efficiency gains: 91.9–96.6\% token reductions on GSM8K (depending on $c$), 76.4–92.9\% on MATH, and 93.1\% on CommonsenseQA.
On GSM8K, $c = 0.98$ reaches 85.50\% accuracy (+31.54pp vs baseline 65.00\%) while using only 1,046 tokens, a 91.9\% reduction from the 12{,}965-token baseline.
On MATH, $c = 0.95$ achieves 83.00\% accuracy (+0.97pp) with 85.4\% fewer tokens.

A key driver of these accuracy gains, especially on GSM8K and CommonsenseQA, is Nemotron's difficulty in producing a final boxed answer within the 13k-token context window.
Many baseline rollouts simply run out of context and are therefore marked incorrect.
By contrast, \lynxmethod{} exits once the calibrated probe deems a cue safe, cutting off these pathological long trajectories and converting them into short, successful question–answer runs.
This allows LYNX to simultaneously reduce tokens by more than 90\% and improve accuracy by large margins relative to the Nemotron baseline on these benchmarks.

These results suggest that Nemotron's verbose generation may stem from a form of overthinking or rambling that LYNX's early-exit mechanism effectively addresses.
By detecting when sufficient reasoning has occurred, LYNX can curtail unhelpful late-stage verbosity, a potential advantage for models that are prone to such patterns.
This also demonstrates LYNX's robustness across architectural families: the same probe training pipeline succeeds on both Qwen and Llama-based models despite their distinct reasoning behaviors.

\section{Additional Experimental Details}
\label{app:exp-details-extended}

\subsection{Token accounting details}

Token counts always refer to \emph{generated} tokens measured under the base model's tokenizer:
\begin{itemize}[leftmargin=12pt,itemsep=1pt]
    \item \textbf{Baseline}: Total tokens in CoT + final answer.
    \item \textbf{DEER}: Total tokens in CoT + final answer, using authors' implementation.
    \item \textbf{ToN}: Total tokens including both visible output and internal tree-search sampling.
    \item \textbf{LYNX}: Total tokens including pre-exit CoT + short answer-only rollout at the exit cue.
\end{itemize}

The ``Tokens'' column in all tables reports mean total tokens per question; $\Delta$Tokens is relative change vs Baseline.

\subsection{Cross-dataset patterns}

\paragraph{Smooth confidence–efficiency tradeoff.}
Higher confidence levels $c$ reliably yield better accuracy preservation at the cost of fewer token savings. As $c$ decreases, LYNX moves smoothly along empirical Pareto frontiers, enabling straightforward deployment-time tuning.

\paragraph{Temperature-dependent behavior.}
At $T=0.0$, LYNX often \emph{improves} accuracy over baseline (particularly GSM8K, MATH-500), suggesting deterministic decoding can drive models into brittle overthinking patterns that early exits regularize by cutting off unhelpful late-chain reasoning. At $T=0.6$, accuracy is typically preserved rather than improved, but token savings remain substantial.

\paragraph{Transfer across tasks and domains.}
Despite training exclusively on Hendrycks MATH, the same probe transfers successfully to GSM8K, AMC 2023, and AIME 2024 (all mathematical but different styles) and CommonsenseQA (non-mathematical), indicating domain-general signals of reasoning correctness rather than MATH-specific memorization.

\paragraph{High-confidence robustness.}
Configurations $c \in \{0.90, 0.95, 0.97\}$ consistently maintain accuracy within 0–2pp of baseline while reducing token usage by 20–50\%, providing strong deployment-ready operating points with both accuracy guarantees and immediate efficiency gains.

\begin{figure*}[t]
\centering
\small
\sffamily 
\definecolor{questionblue}{HTML}{E3F2FD}
\definecolor{baselinered}{HTML}{FFEBEE}
\definecolor{baselinedark}{HTML}{C62828}
\definecolor{lynxorange}{HTML}{C4731C}
\definecolor{lynxlight}{HTML}{FFF3E0}
\definecolor{cueyellow}{HTML}{FFD54F}
\definecolor{explaingreen}{HTML}{E8F5E9}

\begin{tcolorbox}[
    colback=questionblue,
    colframe=questionblue!50!black,
    arc=3mm,
    boxrule=0.5pt,
    left=5pt, right=5pt, top=5pt, bottom=5pt,
    width=\textwidth,
    fontupper=\small\sffamily
]
\textbf{Question:} A new program had 60 downloads in the first month. The number of downloads in the second month was three times as many as the downloads in the first month, but then reduced by 30\% in the third month. How many downloads did the program have total over the three months?
\end{tcolorbox}

\vspace{2mm}

\begin{minipage}[t]{0.48\textwidth}
\vspace{0pt}%
\begin{tcolorbox}[
    colback=baselinered,
    colframe=baselinered!50!black,
    arc=3mm,
    boxrule=0.5pt,
    left=4pt, right=4pt, top=4pt, bottom=4pt,
    fontupper=\footnotesize\sffamily,
    title={\textbf{Baseline (Full Generation)}},
    coltitle=white,
    colbacktitle=baselinedark,
    height=19cm
]
Okay, so I need to figure out the total number of downloads for this program over three months. Let me read the problem again and break it down step by step.

First month: 60 downloads. That seems straightforward.

Then, the second month's downloads were three times as many as the first month. \colorbox{cueyellow}{\textbf{Hmm}}, okay, so if the first month was 60, then the second month would be 3 times 60. Let me write that down to keep track.

First month downloads: 60.

Second month downloads: 3 * 60. Hmm, 3 times 60 is... 180. So, 180 downloads in the second month.

\colorbox{cueyellow}{\textbf{Wait}}, but then the third month description says that the downloads were reduced by 30\%. Okay, so the third month's downloads are 30\% less than the second month. So, first, I need to find out what 30\% of 180 is, and then subtract that from 180 to get the downloads for the third month. Let me calculate that.

30\% of 180: To find 30\%, I can multiply 180 by 0.3. So, 180 * 0.3. Let me do that math. 180 divided by 10 is 18, so 0.3 is 30\% of 180. So, 180 - 54 is 126. Wait, hold on, 180 - 30\% of 180. So, 180 - 54 is 126. So, 126 downloads in the third month.

Now, I need to find the total downloads over the three months. So, I have to add up the downloads from each month. That would be first month, second month, and third month.

So, that's 60 (first month) + 180 (second month) + 126 (third month). Let me add them step by step.

First, 60 + 180 is... 240. Then, 240 + 126. Let me do that: 240 + 100 is 340, then plus 26 is 366. So, total downloads over three months would be 366.

\vspace{1mm}
\centering\textcolor{gray}{\textit{... (193 more tokens of repeated checking and rephrasing) ...}}
\vspace{1mm}

\raggedright
\colorbox{cueyellow}{\textbf{Wait}}, let me double-check my calculations because that seems a bit high. Maybe I made a mistake in the third month's downloads. Let me go over that again.

Second month: 3 * 60 = 180. Third month: 180 reduced by 30\%. So, 30\% of 180 is 54, so 180 - 54 is 126. So, that part seems correct.

Adding 60 + 180 + 126. Let me add them another way to verify. 60 + 180 is 240, and 240 + 126. \colorbox{cueyellow}{\textbf{Hmm}}, 240 + 100 is 340, and 340 + 26 is 366. That still seems correct.

\vfill

\textbf{Final Answer:} $\boxed{366}$

\vspace{2mm}
\raggedleft\textcolor{gray}{\footnotesize Total tokens: \textbf{875}}
\end{tcolorbox}
\end{minipage}
\hfill
\begin{minipage}[t]{0.48\textwidth}
\vspace{0pt}%
\begin{tcolorbox}[
    colback=lynxlight,
    colframe=lynxorange,
    arc=3mm,
    boxrule=0.5pt,
    left=4pt, right=4pt, top=4pt, bottom=4pt,
    fontupper=\footnotesize\sffamily,
    title={\textbf{LYNX (Early Exit)}},
    coltitle=white,
    colbacktitle=lynxorange,
    height=19cm
]
Okay, so I need to figure out the total number of downloads for this program over three months. Let me read the problem again and break it down step by step.

First month: 60 downloads. That seems straightforward.

Then, the second month's downloads were three times as many as the first month. \colorbox{cueyellow}{\textbf{Hmm}} \textcolor{lynxorange}{\faTimesCircle} \textcolor{gray}{\footnotesize [Cue 1: $p_{\text{pos}} \approx 0.03$, set = $\{0,1\}$ $\rightarrow$ continue]}, okay, so if the first month was 60, then the second month would be 3 times 60. Let me write that down to keep track.

First month downloads: 60.

Second month downloads: 3 * 60. Hmm, 3 times 60 is... 180. So, 180 downloads in the second month.

\colorbox{cueyellow}{\textbf{Wait}} \textcolor{lynxorange}{\faTimesCircle} \textcolor{gray}{\footnotesize [Cue 2: $p_{\text{pos}} \approx 0.06$, set = $\{0,1\}$ $\rightarrow$ continue]}, but then the third month description says that the downloads were reduced by 30\%. Okay, so the third month's downloads are 30\% less than the second month. 
...

Now, I need to find the total downloads over the three months. So, I have to add up the downloads from each month. That would be first month, second month, and third month.

So, that's 60 (first month) + 180 (second month) + 126 (third month). Let me add them step by step.

First, 60 + 180 is... 240. Then, 240 + 126. Let me do that: 240 + 100 is 340, then plus 26 is 366. So, total downloads over three months would be 366.

\colorbox{cueyellow}{\textbf{Hmm}} \textcolor{green!60!black}{\faCheckCircle} \textcolor{gray}{\footnotesize [Cue 6: $p_{\text{pos}} \approx 1.00$, set = $\{1\}$ at $c=0.95$ $\rightarrow$ \textbf{EXIT}]}

\vfill

\textbf{Final Answer:} $\boxed{366}$

\vspace{2mm}
\raggedleft\textcolor{gray}{\footnotesize Total tokens: \textbf{682} \textcolor{green!60!black}{(\textbf{-22.1\% vs baseline})}}

\vspace{3mm}
\tcblower
\begin{tcolorbox}[
    colback=explaingreen,
    colframe=explaingreen!50!black,
    arc=2mm,
    boxrule=0.3pt,
    left=3pt, right=3pt, top=3pt, bottom=3pt,
    fontupper=\footnotesize\sffamily
]
\textbf{Exit Decision Process (per cue event):}
\begin{itemize}[leftmargin=10pt, itemsep=1pt, topsep=2pt]
    \item \textbf{Cue 1 (``hmm''):} $p_{\text{pos}} \approx 0.03$. Predictive set: $\{0,1\}$ (ambiguous) $\Rightarrow$ \textcolor{lynxorange}{\textbf{no exit}}.
    \item \textbf{Cue 2 (``wait''):} $p_{\text{pos}} \approx 0.06$. Predictive set: $\{0,1\}$ (ambiguous) $\Rightarrow$ \textcolor{lynxorange}{\textbf{no exit}}.
    \\
    ...
    \item \textbf{Cue 4 (``hmm''):} $p_{\text{pos}} \approx 1.00$. Predictive set shrinks to $\{1\}$ at confidence $c = 0.95$ $\Rightarrow$ \textcolor{green!60!black}{\textbf{safe to exit}}.
\end{itemize}
\vspace{1mm}
\textbf{Outcome:} Same correct answer ($\boxed{366}$) as the baseline. LYNX exits when the conformal set becomes singleton and uses 682 tokens versus 875 for the baseline, saving 193 tokens ($\approx 22.1\%$ of the generation) of largely redundant re-checking.
\end{tcolorbox}
\end{tcolorbox}
\end{minipage}
\caption{Example 2 showing LYNX exit behavior on a GSM8K question. The exit is triggered only when the conformal predictive set collapses to $\{1\}$, i.e., when the probe is calibrated to treat the cue as a safe exit at the chosen confidence level.
}
\label{fig:qualitative-example2}
\end{figure*}

This illustrates how the same conformal machinery that underlies the global coverage guarantee also drives the concrete decisions in individual rollouts:
exits are never triggered ``just'' because the probe is confident in a point estimate, but only when the calibrated predictive set deems the cue a safe exit at the chosen confidence level $c = 1 - \delta$.

\begin{figure*}[t]
\centering
\small
\sffamily
\definecolor{questionblue}{HTML}{E3F2FD}
\definecolor{baselinered}{HTML}{FFEBEE}
\definecolor{baselinedark}{HTML}{C62828}
\definecolor{lynxorange}{HTML}{C4731C}
\definecolor{lynxlight}{HTML}{FFF3E0}
\definecolor{cueyellow}{HTML}{FFD54F}
\definecolor{explaingreen}{HTML}{E8F5E9}
\definecolor{errorred}{HTML}{FFCDD2}
\begin{tcolorbox}[
    colback=questionblue,
    colframe=questionblue!50!black,
    arc=3mm,
    boxrule=0.5pt,
    left=5pt, right=5pt, top=5pt, bottom=5pt,
    width=\textwidth,
    fontupper=\small\sffamily
]
\textbf{Question:} There exist real numbers $x$ and $y$, both greater than 1, such that $\log_x(y^x)=\log_y(x^{4y})=10$. Find $xy$.
\end{tcolorbox}
\vspace{2mm}
\begin{minipage}[t]{0.48\textwidth}
\vspace{0pt}%
\begin{tcolorbox}[
    colback=baselinered,
    colframe=baselinedark,
    arc=3mm,
    boxrule=0.5pt,
    left=4pt, right=4pt, top=4pt, bottom=4pt,
    fontupper=\footnotesize\sffamily,
    title={\textbf{Baseline (Context Window Exceeded)}},
    coltitle=white,
    colbacktitle=baselinedark,
    height=18cm
]
Okay, so I have this problem here: there are real numbers $x$ and $y$, both greater than 1, such that $\log_x(y^x) = 10$ and $\log_y(x^{4y}) = 10$. I need to find the product $xy$. \colorbox{cueyellow}{\textbf{Hmm}}, let me try to figure this out step by step.

First, I remember that logarithms can be tricky, but maybe I can use some logarithm properties to simplify these equations. The problem gives me two equations: [...]

I think I can use the power rule of logarithms here, which says that $\log_b(a^c) = c \cdot \log_b(a)$. Let me apply that to both equations.

Starting with the first equation: $\log_x(y^x) = x \cdot \log_x(y) = 10$. So, that simplifies to $x \cdot \log_x(y) = 10$.

Similarly, the second equation: $\log_y(x^{4y}) = 4y \cdot \log_y(x) = 10$. So, that becomes $4y \cdot \log_y(x) = 10$.

Now, I have two equations: (1) $x \cdot \log_x(y) = 10$, and (2) $4y \cdot \log_y(x) = 10$.

\colorbox{cueyellow}{\textbf{Hmm}}, I notice that both equations involve $\log_x(y)$ and $\log_y(x)$. Maybe I can express one in terms of the other. I remember that $\log_b(a)$ is equal to $1 / \log_a(b)$. So, $\log_x(y) = 1 / \log_y(x)$. Let me denote $\log_y(x)$ as some variable to make it easier. Let's say: Let's let $a = \log_y(x)$. Then, $\log_x(y) = 1/a$.

Substituting back into the equations: (1) $x \cdot (1/a) = 10$, and (2) $4y \cdot a = 10$.

So, equation 1 becomes $x/a = 10$, which means $x = 10a$. Equation 2 becomes $4y \cdot a = 10$, which simplifies to $y \cdot a = 10/4 = 2.5$, so $y = 2.5/a$.

\vspace{1mm}
\centering\textcolor{gray}{\textit{... (5,300 more tokens attempting algebraic manipulations)}}
\vspace{1mm}
\raggedright

\colorbox{cueyellow}{\textbf{Wait}}, maybe I can try numerical methods. Let me define $f(a) = a \cdot \ln(2.5) - \ln(a) \cdot (a+1) - \ln(10)$ and try to find where $f(a) = 0$.

\vspace{1mm}
\centering\textcolor{gray}{\textit{... (8,200 more tokens of repetitive numerical trials and circular reasoning) ...}}
\vspace{1mm}
\raggedright

\colorbox{cueyellow}{\textbf{Wait}}, maybe I can use the fact that $x$ and $y$ are expressed in terms of $k$ [...]

\vfill
\begin{tcolorbox}[
    colback=errorred,
    colframe=errorred!50!black,
    arc=2mm,
    boxrule=0.3pt,
    left=3pt, right=3pt, top=3pt, bottom=3pt,
    fontupper=\footnotesize\sffamily
]
\textbf{\textcolor{red!70!black}{GENERATION TERMINATED}}

Model exceeded 16k token context window limit during generation. No final answer produced.
\end{tcolorbox}
\vspace{2mm}
\raggedleft\textcolor{gray}{\footnotesize Total tokens: \textbf{16,000 (FAILED)}}
\end{tcolorbox}
\end{minipage}
\hfill
\begin{minipage}[t]{0.48\textwidth}
\vspace{0pt}%
\begin{tcolorbox}[
    colback=lynxlight,
    colframe=lynxorange,
    arc=3mm,
    boxrule=0.5pt,
    left=4pt, right=4pt, top=4pt, bottom=4pt,
    fontupper=\footnotesize\sffamily,
    title={\textbf{LYNX (Early Exit)}},
    coltitle=white,
    colbacktitle=lynxorange,
    height=18cm
]
Okay, so I have this problem here: there are real numbers $x$ and $y$, both greater than 1, such that $\log_x(y^x) = 10$ and $\log_y(x^{4y}) = 10$. I need to find the product $xy$. \colorbox{cueyellow}{\textbf{Hmm}} \textcolor{lynxorange}{\faTimesCircle} \textcolor{gray}{\footnotesize [Cue 1: $p_{\text{pos}} \approx 0.04$, set = $\{0,1\}$ $\rightarrow$ continue]}, let me try to figure this out step by step.

First, I remember that logarithms can be tricky, but maybe I can use some logarithm properties to simplify these equations. [...]

Starting with the first equation: $\log_x(y^x) = x \cdot \log_x(y) = 10$. So, that simplifies to $x \cdot \log_x(y) = 10$.

Similarly, the second equation: $\log_y(x^{4y}) = 4y \cdot \log_y(x) = 10$. So, that becomes $4y \cdot \log_y(x) = 10$.

\colorbox{cueyellow}{\textbf{Hmm}} \textcolor{lynxorange}{\faTimesCircle} \textcolor{gray}{\footnotesize [Cue 2: $p_{\text{pos}} \approx 0.10$, set = $\{0,1\}$ $\rightarrow$ continue]}, I notice that both equations involve $\log_x(y)$ and $\log_y(x)$. Maybe I can express one in terms of the other. I remember that $\log_b(a) = 1 / \log_a(b)$. So, $\log_x(y) = 1 / \log_y(x)$.

Let me denote $\log_y(x)$ as $k$. Then, $\log_x(y) = 1/k$.

Substituting back: $x \cdot (1/k) = 10 \Rightarrow x = 10k$, and $4y \cdot k = 10 \Rightarrow y = 10/(4k) = 5/(2k)$.

\vspace{1mm}
\centering\textcolor{gray}{\textit{... (2,100 more tokens with focused reasoning) ...}}
\vspace{1mm}
\raggedright

Now, I can compute $xy$: $xy = 10k \cdot \frac{5}{2k} = \frac{10k \cdot 5}{2k} = \frac{50k}{2k} = \frac{50}{2} = 25$.

\colorbox{cueyellow}{\textbf{Wait}} \textcolor{green!60!black}{\faCheckCircle} \textcolor{gray}{\footnotesize [Cue 24: $p_{\text{pos}} \approx 0.999$, set = $\{1\}$ at $c=0.80$ $\rightarrow$ \textbf{EXIT}]}

\vfill
\textbf{Final Answer:} $\boxed{25}$
\vspace{2mm}
\raggedleft\textcolor{gray}{\footnotesize Total tokens: \textbf{2,543} \textcolor{green!60!black}{(\textbf{84.1\% saved vs. 16k limit})}}
\vspace{3mm}
\tcblower
\begin{tcolorbox}[
    colback=explaingreen,
    colframe=explaingreen!50!black,
    arc=2mm,
    boxrule=0.3pt,
    left=3pt, right=3pt, top=3pt, bottom=3pt,
    fontupper=\footnotesize\sffamily
]
\textbf{Critical Advantage of Early Exit:}
\begin{itemize}[leftmargin=10pt, itemsep=1pt, topsep=2pt]
    \item Baseline enters circular reasoning loops, exceeding the 16k token context window without an answer.
    \item LYNX monitors 24 natural reasoning cues throughout generation.
    \item At Cue 24 ($p_{\text{pos}} \approx 0.999$), conformal set collapses to $\{1\}$ at $c=0.80$ $\Rightarrow$ \textcolor{green!60!black}{\textbf{confident exit}}.
\end{itemize}
\vspace{1mm}
\textbf{Outcome:} Baseline \textit{fails completely}, while LYNX produces correct answer ($\boxed{25}$) using only 2,543 tokens—demonstrating that early exit enables solving problems that would otherwise exceed context limits.
\end{tcolorbox}
\end{tcolorbox}
\end{minipage}
\caption{Qualitative example demonstrating LYNX's critical advantage on a challenging MATH problem. The baseline model enters circular reasoning and exceeds the 16k token context window without producing an answer. In contrast, LYNX monitors 24 natural reasoning cues (\colorbox{cueyellow}{hmm}, \colorbox{cueyellow}{wait}) and exits confidently when the conformal predictive set collapses to $\{1\}$ at the 24th cue. LYNX produces the correct answer using only 2,543 tokens (84.1\% reduction vs. context limit), demonstrating that early exit enables solving problems that would otherwise fail due to context constraints.
}
\label{fig:qualitative-example-context}
\end{figure*}

\begin{figure*}[t]
\centering
\small
\sffamily

\definecolor{questionblue}{HTML}{E3F2FD}
\definecolor{baselinered}{HTML}{FFEBEE}
\definecolor{baselinedark}{HTML}{C62828}
\definecolor{lynxorange}{HTML}{C4731C}
\definecolor{lynxlight}{HTML}{FFF3E0}
\definecolor{cueyellow}{HTML}{FFD54F}
\definecolor{explaingreen}{HTML}{E8F5E9}

\begin{tcolorbox}[
    colback=questionblue,
    colframe=questionblue!50!black,
    arc=3mm,
    boxrule=0.5pt,
    left=5pt, right=5pt, top=5pt, bottom=5pt,
    width=\textwidth,
    fontupper=\small\sffamily
]
\textbf{Question:} Dolly has two books. Pandora has one. If both Dolly and Pandora read each others' books as well as their own, how many books will they collectively read by the end?
\end{tcolorbox}

\vspace{2mm}

\begin{minipage}[t]{0.48\textwidth}
\begin{tcolorbox}[
    colback=baselinered,
    colframe=baselinered!50!black,
    arc=3mm,
    boxrule=0.5pt,
    left=4pt, right=4pt, top=4pt, bottom=4pt,
    fontupper=\footnotesize\sffamily,
    title={\textbf{Baseline (Overthinking \& Error)}},
    coltitle=white,
    colbacktitle=baselinedark,
    height=17.5cm
]
Okay, so I just came across this problem, and I want to figure it out step by step. Let me read it again: Dolly has two books, and Pandora has one. Both of them read each other's books as well as their own. The question is, how many books will they collectively read by the end?

\colorbox{cueyellow}{\textbf{Hmm}}, alright. Let's break this down. First, I need to understand exactly what's being asked. It's about counting the total number of books that Dolly and Pandora will read when they each read each other's books and their own.

So, Dolly has two books. Let me call them Book A and Book B. Pandora has one book, which I can call Book C.

Now, both of them will read each other's books. That means Dolly will read both of Pandora's books and her own, right? So, Dolly has to read Book C (Pandora's book) and Book A and Book B (her own). Similarly, Pandora will read both of Dolly's books and her own. So, Pandora will read Book A and Book B, as well as Book C.

\colorbox{cueyellow}{\textbf{Wait}}, so let me clarify this. Dolly has two books: A and B. She also has one book: C. So, when Dolly reads each other's books, she reads A, B, and C. Similarly, Pandora has books A and B, and reads C as well.

Therefore, each of them is reading all three books. So, Dolly reads three books, and Pandora reads three books. So, together, how many books are they reading? Well, if each reads three books, that would seem like six books in total.

\colorbox{cueyellow}{\textbf{But hold on}}, let me make sure I'm not double-counting. Each book is being read by both Dolly and Pandora, right? So, Book A is read by both, Book B is read by both, and Book C is read by both. So, even though each of them is reading all three books, the total number of books they collectively read is still three...

\vspace{1mm}
\centering\textcolor{gray}{\textit{... (325 more tokens of confusion between "distinct" vs "collective" counts) ...}}
\vspace{1mm}

\raggedright
So, the total distinct books are just three. So, the answer is three.

\vfill

\textbf{Final Answer:} $\boxed{3}$ \textcolor{baselinedark}{\textbf{(Incorrect)}}

\vspace{2mm}
\raggedleft\textcolor{gray}{\footnotesize Total tokens: \textbf{762}}
\end{tcolorbox}
\end{minipage}
\hfill
\begin{minipage}[t]{0.48\textwidth}
\begin{tcolorbox}[
    colback=lynxlight,
    colframe=lynxorange,
    arc=3mm,
    boxrule=0.5pt,
    left=4pt, right=4pt, top=4pt, bottom=4pt,
    fontupper=\footnotesize\sffamily,
    title={\textbf{LYNX (Early Exit)}},
    coltitle=white,
    colbacktitle=lynxorange,
    height=17.5cm
]
Okay, so I just came across this problem, and I want to figure it out step by step. Let me read it again: Dolly has two books, and Pandora has one. Both of them read each other's books as well as their own. The question is, how many books will they collectively read by the end?

\colorbox{cueyellow}{\textbf{Hmm}} \textcolor{lynxorange}{\faTimesCircle} \textcolor{gray}{\footnotesize [Cue 1: $p_{\text{pos}} \approx 0.58$, set = $\{0,1\}$ $\rightarrow$ continue]}, alright. Let's break this down. First, I need to understand exactly what's being asked. It's about counting the total number of books that Dolly and Pandora will read when they each read each other's books and their own.

So, Dolly has two books. Let me call them Book A and Book B. Pandora has one book, which I can call Book C.

Now, both of them will read each other's books. That means Dolly will read both of Pandora's books and her own, right? So, Dolly has to read Book C (Pandora's book) and Book A and Book B (her own). Similarly, Pandora will read both of Dolly's books and her own. So, Pandora will read Book A and Book B, as well as Book C.

\colorbox{cueyellow}{\textbf{Wait}} \textcolor{green!60!black}{\faCheckCircle} \textcolor{gray}{\footnotesize [Cue 2: $p_{\text{pos}} \approx 1.00$, set = $\{1\}$ at $c=0.80$ $\rightarrow$ \textbf{EXIT}]}

\vfill

\textbf{Final Answer:} $\boxed{6}$ \textcolor{green!60!black}{\textbf{(Correct)}}

\vspace{2mm}
\raggedleft\textcolor{gray}{\footnotesize Total tokens: \textbf{377} \textcolor{green!60!black}{(\textbf{-50.5\% vs baseline})}}

\vspace{3mm}
\tcblower
\begin{tcolorbox}[
    colback=explaingreen,
    colframe=explaingreen!50!black,
    arc=2mm,
    boxrule=0.3pt,
    left=3pt, right=3pt, top=3pt, bottom=3pt,
    fontupper=\footnotesize\sffamily
]
\textbf{Preventing Overthinking:}
\begin{itemize}[leftmargin=10pt, itemsep=1pt, topsep=2pt]
    \item \textbf{Baseline Failure:} The model correctly calculates 3 books per person but then overthinks the word "collectively," confuses it with "distinct" books, and hallucinates an incorrect answer ($\boxed{3}$).
    \item \textbf{LYNX Success:} At \textbf{Cue 2 (``Wait'')}, the probe detects that the current state (Dolly: 3, Pandora: 3) is sufficient to answer the question correctly.
    \item \textbf{Outcome:} LYNX exits \textit{before} the model spirals into the "distinct books" confusion, achieving the correct answer ($\boxed{6}$) and saving 50\% of the compute.
\end{itemize}
\end{tcolorbox}
\end{tcolorbox}
\end{minipage}

\caption{Example illustrating how LYNX prevents "overthinking" on a logic puzzle. The baseline model generates sound reasoning initially but triggers a "Wait" cue and subsequently talks itself out of the correct answer, confusing "collective reads" with "distinct books." LYNX monitors the generation, detects that the internal state at the "Wait" cue ($p_{\text{pos}} \approx 1.00$) is robust enough to exit, and produces the correct answer ($\boxed{6}$) while determining that the subsequent reasoning (which led to the error in the baseline) was unnecessary.}
\label{fig:qualitative-example-books}
\end{figure*}

\begin{table*}[t]
\centering
\caption{
Accuracy--efficiency tradeoffs at temperature $T = 0.6$ for DeepSeek-R1-Distill-Qwen-1.5B and QwQ-32B.
For each dataset we report vanilla chain-of-thought (\textsc{Baseline}), DEER, Think-or-Not (ToN), and \lynxmethod{} for multiple confidence levels $c$.
$\Delta$Acc vs base and $\Delta$Tokens vs base are relative changes (\%) with respect to each model's own \textsc{Baseline}.
}
\label{tab:main_t06_two_models}
\scriptsize
\setlength{\tabcolsep}{3pt}
\renewcommand{\arraystretch}{1.05}
\begin{tabular}{l l c
        !{\color{gray}\vrule width 0.8pt}
        cc cc
        !{\color{gray}\vrule width 0.8pt}
        cc cc}
\toprule
 &  &  &
\multicolumn{4}{c}{DeepSeek-R1-1.5B} &
\multicolumn{4}{c}{QwQ-32B} \\
\cmidrule(lr){4-7} \cmidrule(lr){8-11}
Dataset & Method & Conf &
Acc & $\Delta$ vs base & Avg. & $\Delta$ vs base &
Acc & $\Delta$ vs base & Avg. & $\Delta$ vs base \\
\midrule
\multirow{8}{*}{\textsc{GSM8K}} 
  & \textsc{Baseline}      & --   & 0.7839 &   -- & 1036.79 &   -- & 0.9500 &   -- & 1450.68 &   -- \\
\cmidrule(lr){2-11}
  & \textsc{DEER}          & --   & 0.6755 & -13.83\% &  649.49 & -37.4\%  & \textbf{0.9439} &  \textbf{-0.64\%} &  \textbf{983.77} & \textbf{-32.2\%} \\
  & \textsc{ToN}           & --   & 0.6641 & -15.28\% &  751.75 & -27.5\%  & 0.8104 & -14.69\% &  1194.89 & -17.6\% \\
\cmidrule(lr){2-11}
  & \lynxmethod{}          & 0.97 & 0.7832 &  -0.09\% &  832.48 & -19.7\%  & 0.9356 &  -1.52\% & 1095.10 & -24.5\% \\
  & \lynxmethod{}          & 0.95 & 0.7817 &  -0.28\% &  721.28 & -30.4\%  & 0.9249 &  -2.64\% &  974.08 & -32.9\% \\
  & \textbf{\lynxmethod{}} & \textbf{0.90} & \textbf{0.7786} & \textbf{-0.68\%} & \textbf{592.45} & \textbf{-42.9\%} & 0.8999 &  -5.27\% &  808.08 & -44.3\% \\
  & \lynxmethod{}          & 0.80 & 0.7460 &  -4.83\% &  518.81 & -50.0\%  & 0.8249 & -13.17\% &  603.34 & -58.4\% \\
  & \lynxmethod{}          & 0.70 & 0.7036 & -10.24\% &  472.69 & -54.4\%  & 0.6839 & -28.01\% &  451.89 & -68.8\% \\
\midrule
\multirow{8}{*}{\textsc{MATH-500}} 
  & \textsc{Baseline}      & --   & 0.8180 &   -- & 4598.35 &   -- & 0.9100 &   -- & 3895.07 &   -- \\
\cmidrule(lr){2-11}
  & \textsc{DEER}          & --   & 0.6800 & -16.87\% & 2701.16 & -41.3\%  & \textbf{0.9160} &  \textbf{+0.66\%} & \textbf{3039.63} & \textbf{-22.0\%} \\
  & \textsc{ToN}           & --   & 0.5980 & -26.89\% & 3159.85 & -31.3\%  & 0.7160 & -21.32\% &  2517.44 & -35.4\% \\
\cmidrule(lr){2-11}
  & \lynxmethod{}          & 0.97 & 0.8060 &  -1.47\% & 3599.51 & -21.7\%  & 0.8920 &  -1.98\% & 3067.14 & -21.3\% \\
  & \textbf{\lynxmethod{}} & \textbf{0.95} & \textbf{0.8000} & \textbf{-2.20\%} & \textbf{3128.06} & \textbf{-32.0\%} & 0.8780 &  -3.52\% & 2811.12 & -27.8\% \\
  & \lynxmethod{}          & 0.90 & 0.7700 &  -5.87\% & 2600.05 & -43.5\%  & 0.8400 &  -7.69\% & 2443.15 & -37.3\% \\
  & \lynxmethod{}          & 0.80 & 0.6800 & -16.87\% & 1866.73 & -59.4\%  & 0.7920 & -12.97\% & 1925.46 & -50.6\% \\
  & \lynxmethod{}          & 0.70 & 0.5720 & -30.07\% & 1453.57 & -68.4\%  & 0.6600 & -27.47\% & 1365.26 & -64.9\% \\
\midrule
\multirow{8}{*}{\textsc{AMC 2023}} 
  & \textsc{Baseline}      & --   & 0.7500 &   -- & 7118.93 &   -- & 0.9000 &   -- & 6505.08 &   -- \\
\cmidrule(lr){2-11}
  & \textsc{DEER}          & --   & 0.6500 & -13.33\% & 5273.05 & -25.9\%  & 0.9000 &   0.00\% & 6237.00 &  -4.1\% \\
  & \textsc{ToN}           & --   & 0.5000 & -33.33\% & 3875.55 & -45.6\%  & 0.7500 & -16.67\% & 3554.87 & -45.4\% \\
\cmidrule(lr){2-11}
  & \lynxmethod{}          & 0.97 & 0.7500 &   0.00\% & 5725.05 & -19.6\%  & 0.8750 &  -2.78\% & 4817.25 & -25.9\% \\
  & \lynxmethod{}          & 0.95 & 0.7000 &  -6.67\% & 4754.60 & -33.2\%  & \textbf{0.8500} &  \textbf{-5.56\%} & \textbf{4551.03} & \textbf{-30.0\%} \\
  & \textbf{\lynxmethod{}} & \textbf{0.90} & \textbf{0.7250} &  \textbf{-3.33\%} & \textbf{3423.63} & \textbf{-51.9\%} & 0.7750 & -13.89\% & 4042.70 & -37.9\% \\
  & \lynxmethod{}          & 0.80 & 0.6250 & -16.67\% & 2543.48 & -64.3\%  & 0.6750 & -25.00\% & 3144.23 & -51.7\% \\
  & \lynxmethod{}          & 0.70 & 0.4250 & -43.33\% & 2070.98 & -70.9\%  & 0.5750 & -36.11\% & 2562.03 & -60.6\% \\
\midrule
\multirow{8}{*}{\textsc{AIME 2024}} 
  & \textsc{Baseline}      & --   & 0.1667 &   -- & 12189.00 &   -- & 0.6000 &   -- & 10115.93 &   -- \\
\cmidrule(lr){2-11}
  & \textsc{DEER}          & --   & \textbf{0.3000} & \textbf{+79.96\%} &  \textbf{8303.06} & \textbf{-31.9\%}  & 0.6667 & +11.12\% &  8767.23 & -13.3\% \\
  & \textsc{ToN}           & --   & 0.2000 & +19.98\% & 10408.63 & -14.6\%  & 0.5000 & -16.67\% &  8111.83 & -19.8\% \\
\cmidrule(lr){2-11}
  & \lynxmethod{}          & 0.97 & 0.1667 &   0.00\% & 10817.13 & -11.3\%  & 0.6000 &   0.00\% &  7343.97 & -27.4\% \\
  & \textbf{\lynxmethod{}} & \textbf{0.95} & \textbf{0.1667} & \textbf{0.00\%} & \textbf{9963.00} & \textbf{-18.3\%} & \textbf{0.6000} &   \textbf{0.00\%} &  \textbf{6743.93} & \textbf{-33.3\%} \\
  & \lynxmethod{}          & 0.90 & 0.1333 & -20.04\% & 7814.43 & -35.9\%  & 0.5000 & -16.67\% &  5940.80 & -41.3\% \\
  & \lynxmethod{}          & 0.80 & 0.1333 & -20.04\% & 5503.27 & -54.9\%  & 0.4667 & -22.22\% &  5262.70 & -48.0\% \\
  & \lynxmethod{}          & 0.70 & 0.1000 & -40.01\% & 3900.03 & -68.0\%  & 0.3333 & -44.45\% &  4248.90 & -58.0\% \\
\midrule
\multirow{8}{*}{%
  \begin{tabular}{@{}l@{}}
    \textsc{CommonsenseQA} \\
    (Non-Math)
  \end{tabular}
}
  & \textsc{Baseline}      & --   & 0.4218 &   -- &  803.10 &   -- & 0.8501 &   -- &  950.51 &   -- \\
\cmidrule(lr){2-11}
  & \textsc{DEER}          & --   & 0.4275 &  +1.35\% &  590.80 & -26.4\%  & 0.8477 &  -0.28\% &  643.47 & -32.3\% \\
  & \textsc{ToN}           & --   & 0.4529 &  +7.37\% &  559.95 & -30.3\%  & \textbf{0.8706} &  \textbf{+2.41\%} &    \textbf{607.19} & \textbf{-36.1\%} \\
\cmidrule(lr){2-11}
  & \lynxmethod{}          & 0.97 & 0.4226 &  +0.19\% &  752.80 &  -6.3\%  & 0.8567 &  +0.78\% &  733.78 & -22.8\% \\
  & \lynxmethod{}          & 0.95 & 0.4242 &  +0.57\% &  738.90 &  -8.0\%  & 0.8518 &  +0.20\% &  653.96 & -31.2\% \\
  & \lynxmethod{}          & 0.90 & 0.4292 &  +1.75\% &  691.10 & -13.9\%  & 0.8501 &   0.00\% &  541.47 & -43.0\% \\
  & \lynxmethod{}          & 0.80 & 0.4324 &  +2.51\% &  599.90 & -25.3\%  & 0.8337 &  -1.93\% &  402.35 & -57.7\% \\
  & \textbf{\lynxmethod{}} & \textbf{0.70} & \textbf{0.4210} & \textbf{-0.19\%} & \textbf{365.00} & \textbf{-54.6\%} & \textbf{0.8346} &  \textbf{-1.82\%} &  \textbf{331.83} & \textbf{-65.1\%} \\
\bottomrule
\end{tabular}
\end{table*}

\end{document}